\documentclass[journal]{template/IEEEtran}

\usepackage{bmpsize}
\usepackage{amsthm}
\usepackage[linesnumbered,ruled,vlined]{algorithm2e}
\usepackage{amsmath,amssymb,enumerate}
\usepackage{mathrsfs}
\usepackage{times}
\usepackage{multicol}
\usepackage{amsfonts}
\usepackage{textcomp}
\usepackage{balance}
\usepackage{multirow}
\usepackage{booktabs}
\usepackage{dblfloatfix} 
\usepackage{bbold}
\usepackage{makecell}
\usepackage{hhline} 
\usepackage{float} 
\usepackage{xparse}
\usepackage{lipsum}
\usepackage{soul} 
\usepackage{times}
\usepackage{cite}
\usepackage{balance}
\usepackage{dsfont} 
\usepackage{physics} 
\usepackage{setspace} 
\usepackage{accents} 

\usepackage[dvipdfmx]{graphicx} 
\usepackage[T1]{fontenc}
\usepackage[usenames,dvipsnames,svgnames,table, xcdraw]{xcolor}
\usepackage[utf8]{inputenc}

\usepackage{hyperref}
\hypersetup{
  colorlinks=true,pdfpagemode=UseNone,citecolor=black,linkcolor=black,urlcolor=BrickRed
}

\usepackage{caption}
\usepackage{cleveref}
\usepackage{subcaption}

\SetCommentSty{mycommfont}

\SetKwInput{KwInput}{Input}                
\SetKwInput{KwOutput}{Output}              

\graphicspath{
    {graphics/}
}

\newtheoremstyle{break}
  {\topsep}{\topsep}%
  {\itshape}{}%
  {\bfseries}{}%
  {\newline}{}%

\newtheorem{remark}{Remark}

\theoremstyle{break}

\definecolor{note}{rgb}{0.3,0.7,0.25}
\definecolor{rephase}{rgb}{0.15,0.7,0.15}
\definecolor{bag}{rgb}{0.5,0.5,0.0}

\newcommand{\lidar}{LiDAR~}

\newcommand{\xaxisN}{$x$-axis}
\newcommand{\yaxisN}{$y$-axis}

\newcommand{\xaxis}{$x$-axis~}
\newcommand{\yaxis}{$y$-axis~}

\newcommand{\AstarN}{$\text{A}^*$}

\newcommand{\rrt}{$\text{RRT}^*$~}
\newcommand{\rrtN}{$\text{RRT}^*$}

\newcommand{\xfree}{\Xcal_{\text{free}}}
\newcommand{\traj}{\mathscr{T}}
\newcommand{\nsub}[1]{n_{\text{#1}}}
\newcommand{\csub}[1]{c_{\text{#1}}}

\newcommand{\comment}[1]{}

\def\reals{\mathbb{R}}

\DeclareDocumentCommand{\RI}{ O{} O{} }{\mathcal{RI}_{#1}^{#2}}

\DeclareDocumentCommand{\td}{ O{} }{\tilde{#1}}

\DeclareDocumentCommand{\asin}{ O{} }{\sin^{-1}(#1)}
\DeclareDocumentCommand{\acos}{ O{} }{\cos^{-1}(#1)}
\DeclareDocumentCommand{\atan}{ O{} }{\tan^{-1}(#1)}

\DeclareDocumentCommand{\vector}{ O{} }{\mathrm{vec}(#1)}
\DeclareDocumentCommand{\zeros}{ O{} }{\textbf{0}_{#1}}
\DeclareDocumentCommand{\pre}{ O{} O{} }{{}_{#1}^{#2}}

\DeclareMathOperator*{\argmin}{arg\,min}

\newcommand{\Gcal}{\mathcal{G}}

\newcommand{\Mcal}{\mathcal{M}}
\newcommand{\Ncal}{\mathcal{N}}

\newcommand{\Pcal}{\mathcal{P}}

\newcommand{\Tcal}{\mathcal{T}}

\newcommand{\Xcal}{\mathcal{X}}

\DeclareDocumentCommand{\A}{ O{} O{} }{\textbf{A}_{#1}^{#2}}
\DeclareDocumentCommand{\H}{ O{} O{} }{\textbf{H}_{#1}^{#2}}
\DeclareDocumentCommand{\I}{ O{} O{} }{\textbf{I}_{#1}^{#2}}
\DeclareDocumentCommand{\L}{ O{} O{} }{\textbf{L}_{#1}^{#2}}
\DeclareDocumentCommand{\M}{ O{} O{} }{\textbf{M}_{#1}^{#2}}
\DeclareDocumentCommand{\N}{ O{} O{} }{\textbf{N}_{#1}^{#2}}
\DeclareDocumentCommand{\O}{ O{} O{} }{\textbf{O}_{#1}^{#2}}
\DeclareDocumentCommand{\P}{ O{} O{} }{\textbf{P}_{#1}^{#2}}
\DeclareDocumentCommand{\Q}{ O{} O{} }{\textbf{Q}_{#1}^{#2}}
\DeclareDocumentCommand{\R}{ O{} O{} }{\textbf{R}_{#1}^{#2}}
\DeclareDocumentCommand{\T}{ O{} O{} }{\textbf{T}_{#1}^{#2}}
\DeclareDocumentCommand{\U}{ O{} O{} }{\textbf{U}_{#1}^{#2}}
\DeclareDocumentCommand{\V}{ O{} O{} }{\textbf{V}_{#1}^{#2}}
\DeclareDocumentCommand{\X}{ O{} O{} }{\textbf{X}_{#1}^{#2}}
\DeclareDocumentCommand{\Y}{ O{} O{} }{\textbf{Y}_{#1}^{#2}}
\DeclareDocumentCommand{\Z}{ O{} O{} }{\textbf{Z}_{#1}^{#2}}

\DeclareDocumentCommand{\e}{ O{} O{} }{\textbf{e}_{#1}^{#2}}
\DeclareDocumentCommand{\n}{ O{} O{} }{\textbf{n}_{#1}^{#2}}
\DeclareDocumentCommand{\o}{ O{} O{} }{\textbf{o}_{#1}^{#2}}
\DeclareDocumentCommand{\t}{ O{} O{} }{\textbf{t}_{#1}^{#2}}
\DeclareDocumentCommand{\p}{ O{} O{} }{\textbf{p}_{#1}^{#2}}
\DeclareDocumentCommand{\q}{ O{} O{} }{\textbf{q}_{#1}^{#2}}
\DeclareDocumentCommand{\r}{ O{} O{} }{\textbf{r}_{#1}^{#2}}
\DeclareDocumentCommand{\u}{ O{} O{} }{\textbf{u}_{#1}^{#2}}
\DeclareDocumentCommand{\v}{ O{} O{} }{\textbf{v}_{#1}^{#2}}
\DeclareDocumentCommand{\x}{ O{} O{} }{\textbf{x}_{#1}^{#2}}

\title{Efficient Anytime CLF Reactive Planning System \\for a Bipedal~Robot~on~Undulating~Terrain}

\author{Jiunn-Kai Huang and Jessy W. Grizzle
\thanks{Jiunn-Kai Huang and J. Grizzle, are with the Robotics
Institute, University of Michigan, Ann Arbor, MI 48109, USA. \texttt{\{bjhuang, grizzle\}@umich.edu}.} }

\begin{document}

\maketitle
\pagestyle{plain}

\begin{abstract}
    We propose and experimentally demonstrate a reactive planning system for bipedal
    robots on unexplored, challenging terrains. The system consists of a
    low-frequency planning thread (5 Hz) to find an asymptotically optimal path and a
    high-frequency reactive thread (300 Hz) to accommodate robot deviation. The
    planning thread includes: a multi-layer local map to compute traversability for
    the robot on the terrain; an anytime omnidirectional Control Lyapunov Function
    (CLF) for use with a Rapidly Exploring Random Tree Star (RRT*) that generates a
    vector field for specifying motion between nodes; a sub-goal finder when the
    final goal is outside of the current map; and a finite-state machine to handle
    high-level mission decisions. The system also includes a reactive thread to
    obviate the non-smooth motions that arise with traditional RRT* algorithms when
    performing path following. The reactive thread copes with robot deviation while
    eliminating non-smooth motions via a vector field (defined by a closed-loop
    feedback policy) that provides real-time control commands to the robot's gait
    controller as a function of instantaneous robot pose. The system is evaluated on
    various challenging outdoor terrains and cluttered indoor scenes in both
    simulation and experiment on Cassie Blue, a bipedal robot with 20 degrees of
    freedom. All implementations are coded in C++ with the Robot Operating System
    (ROS) and are available at
    \href{https://github.com/UMich-BipedLab/CLF\_reactive\_planning\_system}{https://github.com/UMich-BipedLab/CLF\_reactive\_planning\_system}.

\end{abstract}


%


\section{Introduction}
\label{sec:Introduction}

\begin{figure}[!t]%
\centering
\begin{subfigure}{0.9\columnwidth}
    \centering
\includegraphics[width=1\columnwidth]{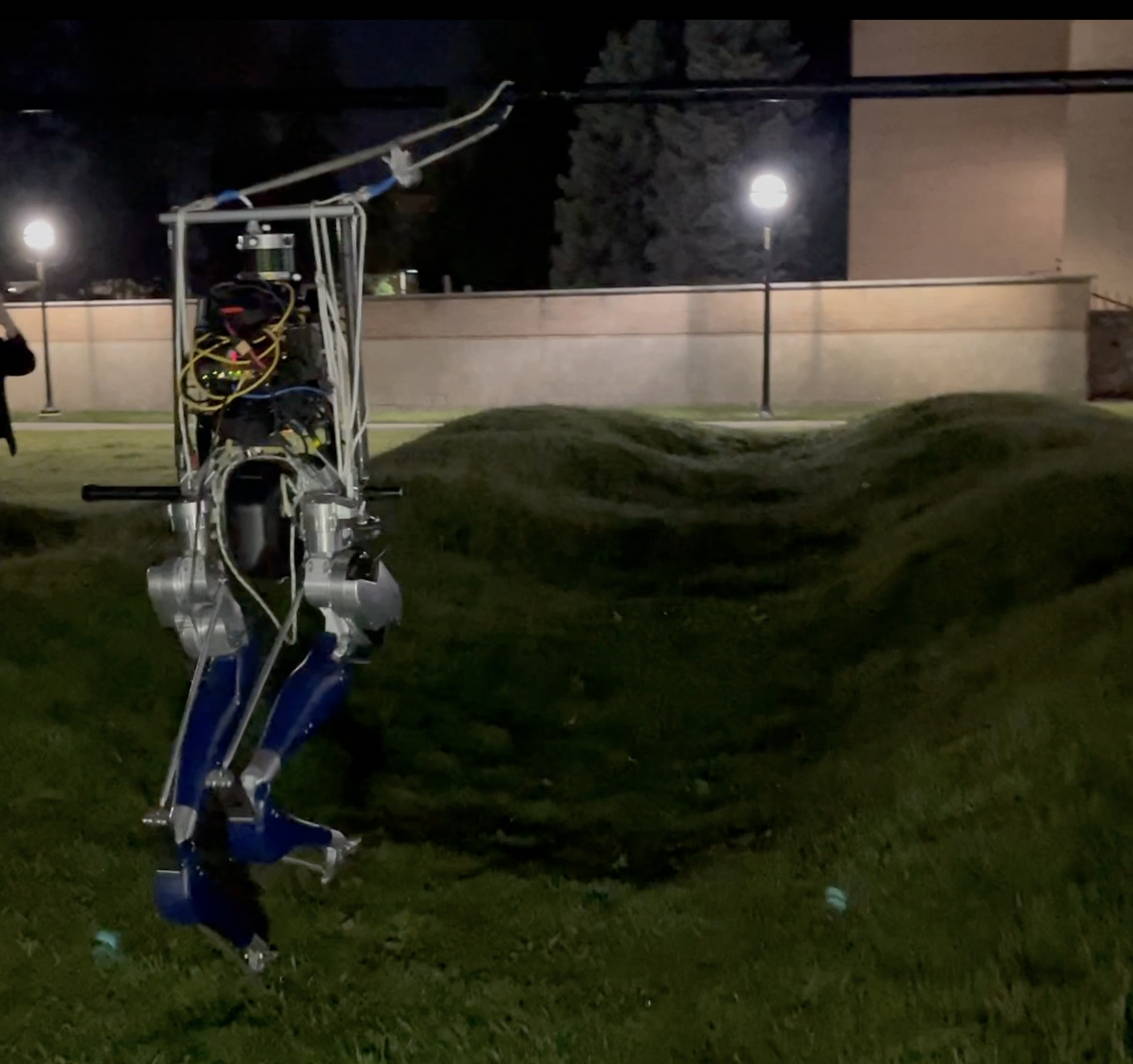}
\end{subfigure}\vspace{5pt}
\begin{subfigure}{0.9\columnwidth}
    \centering
\includegraphics[width=1\columnwidth]{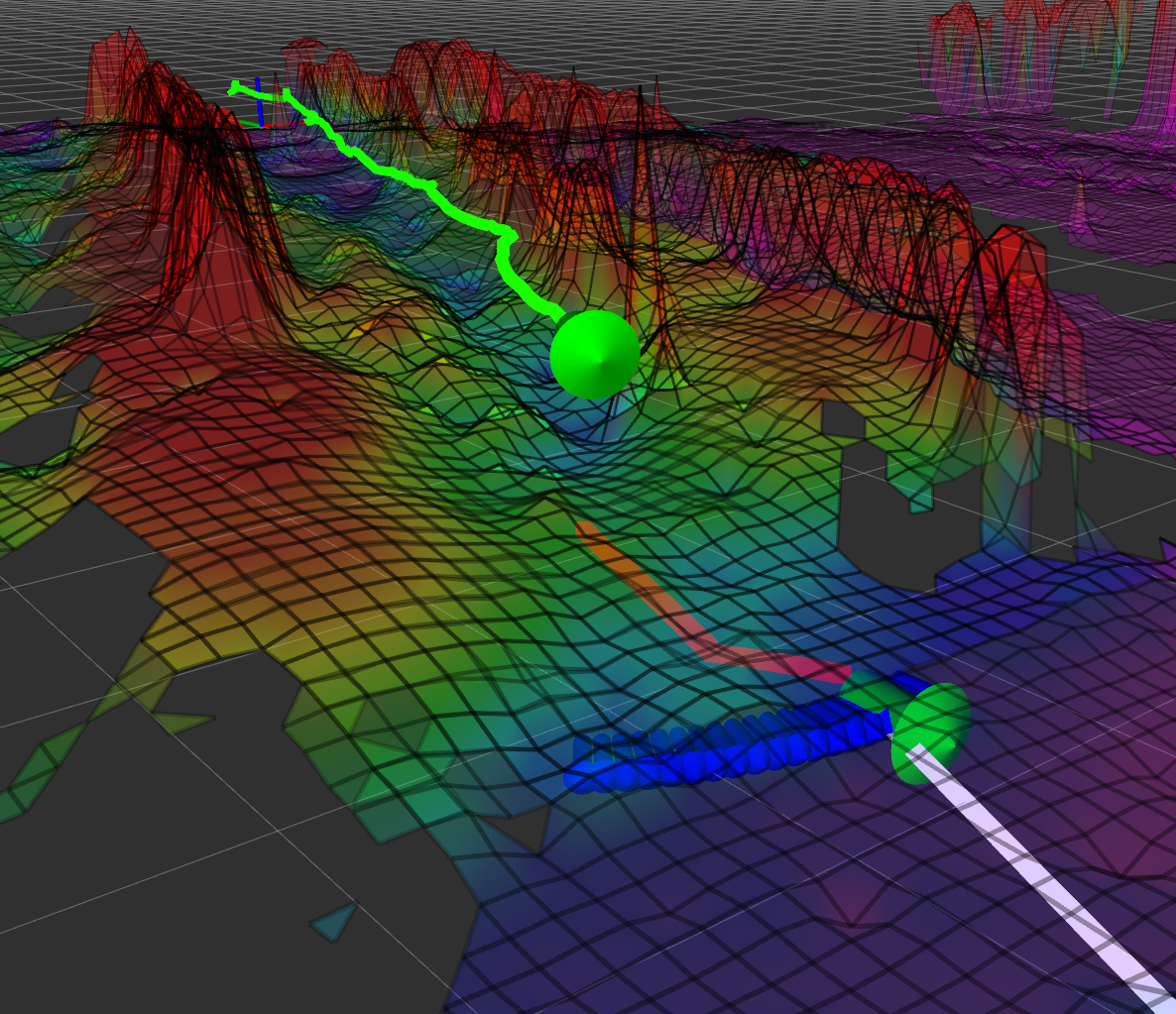}
\end{subfigure}%
\caption[]{In the top figure, Cassie Blue autonomously traverses the Wave Field
    via the proposed reactive planning system, comprised of a planning thread
    and a reactive thread. The planning thread involves a multi-layer local map to
    compute traversability, a sub-goal finder, and an omnidirectional Control Lyapunov Function \rrtN. Instead of a
    common waypoint-following or path-tracking strategy, the reactive thread copes with robot deviation while eliminating non-smooth motions via a vector field (defined by a
    closed-loop feedback policy) that provides real-time control commands to the robot's gait controller as a function of instantaneous robot pose. The bottom figure is the elevation map built online. The red peaks are from the experimenters walking alongside Cassie.}
\label{fig:FirstImage}%
\end{figure}
Motion planning as a central component for autonomous navigation has been extensively studied over the last
few decades. Algorithms such as \rrtN, \AstarN, and their
variants focus on finding an (asymptotically) optimal path as computationally efficiently as
possible\cite{VanillaRRT, lavalle2001randomized, RRTStar, RRTStarICRA,
karaman2010incremental, karaman2010optimal,FastRRT, 7487439, 9082624}. The
application of these algorithms relies on designing a control policy to track the planned path, resulting in waypoint following or pathway tracking.
In turn, the tracking of path segments (between waypoints) leads to non-smooth motion of the actual robot, due to abrupt
acceleration or heading changes when transitioning between waypoints/pathways. 

This paper seeks to develop a reactive planning system for bipedal robots on unexplored, unmapped, challenging terrains and to provide high-rate (directional) velocity and heading commands to be realized by the robot's low-level feedback-control gait-generation algorithm. For this application, the non-smooth aspects of the planned motions arising from waypoints/pathways transitions are detrimental to stability of the overall system.

Several approaches have been developed to address the non-smooth aspects of paths
produced by motion planning, such as reactive motion planning \cite{golbol2018rg,
arslan2019sensor, paternain2017navigation, ExactRobot2016, koditschek1990robot,
rimon1990exact, borenstein1989real, koditschek1987exact, gomez2013planning} and
feedback motion planning  \cite{tedrake2010lqr,Park2DCLF, ParkRRT}.
Fundamentally, these approaches replace paths to be followed with smooth vector
fields whose solutions guide the robot's evolution in its configuration space. 

We are inspired by the work of \cite{Park2DCLF, ParkRRT}, which proposes a
Control Lyapunov Function (CLF) to realize reactive planning for a non-holonomic
differential-drive wheeled robot. While their underlying model is not applicable to a
Cassie bipedal robot, their basic concept is applicable. As part of our work, we
design an appropriate CLF for robots capable of walking in any direction with any
orientation. Moreover, we take into account features specific to bipeds, such as the
limited lateral leg motion that renders lateral walking more laborious than sagittal
plane walking. 


The feedback motion planning algorithm in \cite{Park2DCLF, ParkRRT} has not yet been
evaluated on hardware. In general, there is a significant chasm between a planning
algorithm and autonomous navigation on real robots. Most planning algorithms assume
not only that a fully-explored, noise-free, perfect map is given but also that the
robot's destination will always lie within this map. Moreover, the algorithms also
assume a perfect robot pose and a perfect robot with ideal actuators that can execute
an arbitrary trajectory. These assumptions are not practical. Therefore, utilizing a
planning algorithm for autonomous navigation with real robots remains challenging. We
propose and demonstrate experimentally an autonomous navigation system for a Cassie
bipedal robot that is able to handle a noisy map in real-time, a distant goal that
may not be in the initial map when the user decides where to send the robot, and
importantly, a means to smoothly handle robot deviation. Additionally, a rudimentary
finite-state machine is integrated to handle actions such as where to turn at
intersections.



\section{Related Work and Contributions}
\label{sec:RelatedWork}
Motion planning, an essential component of robot autonomy, has been an active area of research for
multiple decades with an accompanying rich literature. In this section, we review several types of
planning algorithms and summarize our main contributions.

\subsection{Sampling-Based Motion Planning}
\label{sec:SamplingBased}
Rapidly exploring Tree (RRT)\cite{VanillaRRT} stands out for its low complexity and
high efficiency in exploring unknown configuration spaces. Its asymptotically
optimal version --- \rrt\cite{karaman2010incremental} --- has also gained much
attention and has contributed greatly to the spread of the RRT family. RRT, \rrtN,
and variations on the basic algorithms, generate a collision-free path comprised of piece-wise linear
paths between discrete poses of the robot \cite{VanillaRRT, lavalle2001randomized, RRTStar, RRTStarICRA,
karaman2010incremental, karaman2010optimal,FastRRT, 7487439, 9082624,
janson2015fast,otte2016rrtx}. However, abrupt (non-differentiable) transitions between waypoints/pathways
are an inherent issue with this family of planning algorithms and in addition, the generated
trajectories do not account for control constraints. Therefore, to ensure the
produced trajectories are feasible, additional expensive computations such as trajectory
smoothing or optimization are often involved. A great deal of attention has been
directed to this area, resulting in versions of \rrtN
\cite{vailland2021cubic, SmoothRRT, aguilar2017rrt, 7330891,chiang2018colreg,
enevoldsen2021colregs} that utilize different smoothing techniques or steering
functions.

Trajectory smoothing (B-spines, Dubins, or other parametric curves) is often designed
independently of robot dynamics\cite{parque2020smooth, zdevsar2018optimum,
jusko2016scalable}, which can lead to unbounded turning rate, acceleration, or jerk.
Therefore, additional computations are necessary to validate the resulting smoothed
trajectory. Furthermore, these methods are often ambiguous about how they treat robot deviations about the planned path and
in the end provide open-loop control laws for tracking. 


\subsection{Reactive Planning}
Reactive planning contributes another significant concept to the motion
planning literature \cite{golbol2018rg, arslan2019sensor, paternain2017navigation,
ExactRobot2016, koditschek1990robot, rimon1990exact, borenstein1989real,
koditschek1987exact, gomez2013planning}, namely potential fields. The method of reactive planning controls the motion of a
robot by covering the configuration space with a potential field, creating a single
attractive equilibrium around the target point and repulsive actions around obstacles. In
other words, the reactive planning replaces the concept of trajectory with that of a vector field arising as the gradient of a potential
function. The method of potential fields seems to address all the issues raised in
Sec.~\ref{sec:SamplingBased} for sampling-based methods. However, most of the experimental work has been carried out on flat
ground and it is unclear how extensions to undulating terrain can be performed.
Maybe the biggest drawback for real-time applications is that these algorithms require a complete map to construct a potential
field.

The concept of combining sampling-based algorithms with reactive planning was
developed in \cite{tedrake2010lqr,Park2DCLF, ParkRRT}, which not only provides a feasible path to follow
from \rrt, but also a smooth feedback control law that instantaneously replans a path to the next goal as the robot deviates due to model imperfections in the robot's hardware or terrain. The feedback laws greatly ameliorate the issue of non-smooth paths. The feedback motion planning in \cite{tedrake2010lqr} is based on a family of CLFs designed via linearization of the robot's model around a sufficiently large set of points in the robot's state space, LQR, and Sum of Squares (SoS), whereas the
feedback motion planning of \cite{Park2DCLF, ParkRRT} uses a single CLF and varies the associated equilibrium to set sub-goal poses. 

The feedback motion planning of \cite{Park2DCLF, ParkRRT} is a form of non-holonomic \rrt for differential-drive
wheeled robots, where a CLF is utilized as the steering function in the \rrt algorithm
and results in a system where robot control and motion planning are tightly coupled. This method is closest to ours. 
The CLF in \cite{Park2DCLF, ParkRRT} is designed for differential-drive wheeled robots and hence is not suitable for
bipeds, see Sec.~\ref{sec:DeficiencyOfExistingCLF}. Additionally, their version of
\rrt assumes the robot is on flat ground and thus is not appropriate for undulating
terrains. To date, the work has not been evaluated on hardware.

In this paper, we propose a CLF that is suitable for bipeds or omnidirectional robots
(Sec.~\ref{sec:CLF}) and utilize the proposed CLF in the \rrt algorithm; see
Sec.~\ref{Sec:Planner}. We integrate the omnidirectional \rrt into a
full reactive planning system consisting of both a planning thread and a reactive
thread. The planning thread includes a multi-layer, robot-centric local map, an
anytime omnidirectional CLF \rrtN, a sub-goal finder, and a finite-state machine. The
reactive thread utilizes the CLF as a reactive planner to handle robot deviations; see
Sec.~\ref{sec:PlannerSystem}.

\subsection{Contributions}
In particular, the present work has the following contributions:
\begin{enumerate}
    \item We propose a novel smooth Control Lyapunov Function (CLF) with a
        closed-form solution to the feedback controller for omnidirectional robots.
        The CLF is designed such that when a goal is far from the robot position, the
        CLF controls the robot orientation to align with the goal while moving toward
        the goal. On the other hand, the robot
        walks to the goal disregarding its orientation if the goal is close.
        Additionally, we study the behaviors of the CLF under different initial
        conditions and parameters. 
    \item We define a closed-form distance measure from a pose (position and
        orientation) to a target position for omnidirectional robots under a
        pose-centric polar coordinate. This distance metric nicely captures
        inherent features of Cassie-series robots, such as the low-cost of
        longitudinal movement and high-cost of lateral movement.     
    \item We utilize the proposed CLF and the distance measure to form a new variation
        of \rrt (omnidirectional CLF \rrtN) to tackle undulating terrains, in which both
        distance and traversability are included in the cost to solve the optimal path problem.  Moreover, as in \cite{Park2DCLF}, the optimal path is realized as a sequence of subgoals that are connected by integral curves of a set of vector fields, thereby providing reactive planning: in response to a disturbance, each vector field associated with the optimal path automatically guides the robot to a subgoal along a new integral curve of the vector field.

    \item We integrate all the above components together as a reactive planning
        system for challenging terrains/cluttered indoor environments. It contains a
        planning thread to guide Cassie to walk in highly traversable areas toward a
        distant goal on the basis of a multi-layer map being built in real-time and a reactive
        thread to handle robot deviation via a closed-loop feedback control instead
        of a commonly used waypoint-following or path-tracking strategy.
\end{enumerate}

        We evaluate the reactive planning system by performing three types of experiments:
        \textbf{1)} A simplified biped pendulum model (inputs are piece-wise constant,
        similar to Cassie-series robots) navigates various synthetic, noisy,
        challenging outdoor terrains and cluttered indoor scenes. The system guides
        the robot to its goals in various scenes, both indoors and outdoors, 
        with or without obstacles. The system also guides the robot to completion of
        several high-level missions, such as turning left at every
        intersection. 
        \textbf{2)} To verify that the outputs of the control commands are feasible
        for Cassie-series robots, the system gives commands to a Cassie whole-body dynamic
        simulator~\cite{cassiesims}, which simulates 20 degrees of freedom (DoF) of
        Cassie in Matlab Simmechanics on a 3D terrain.
        \textbf{3)} Lastly, the reactive planning system successfully allows Cassie
        Blue to complete several indoor and outdoor missions: a) walking in corridors
        and avoiding furniture in the Ford Robotics Building (FRB) at the University
        of Michigan;
        b) turning left when detected intersections of corridors and return to its
        initial position in FRB; and
        c) traversing parts of the Wave Field on the North campus of the University of Michigan, as shown in Fig~\ref{fig:FirstImage}. 
        
        The videos of the autonomy
        experiments can be found at \cite{WaveFieldAutonomy2021}. All of the simulated environments, the experimental data, and
        the C++ implementations for the reactive planning system are made available at
        \href{https://github.com/UMich-BipedLab/CLF_reactive_planning_system}{https://github.com/UMich-BipedLab/CLF\_reactive\_planning\_system}\cite{githubFileCLFPlanning}.


The remainder of this paper is organized as follows. Section~\ref{sec:CLF} constructs the new CLF 
for bipeds and omnidirectional robots. The omnidirectional CLF \rrt is introduced 
in Sec.~\ref{Sec:Planner}. Section~\ref{sec:PlannerSystem} integrates all the
above components as a reactive planning system. Simulated and experimental evaluations of
the proposed reactive system is presented in Sec.~\ref{sec:Experiment}. Finally,
Sec.~\ref{sec:Conclusion} concludes the paper and provides suggestions for future
work. 

\begin{figure}[t]%
\centering
\includegraphics[width=1\columnwidth]{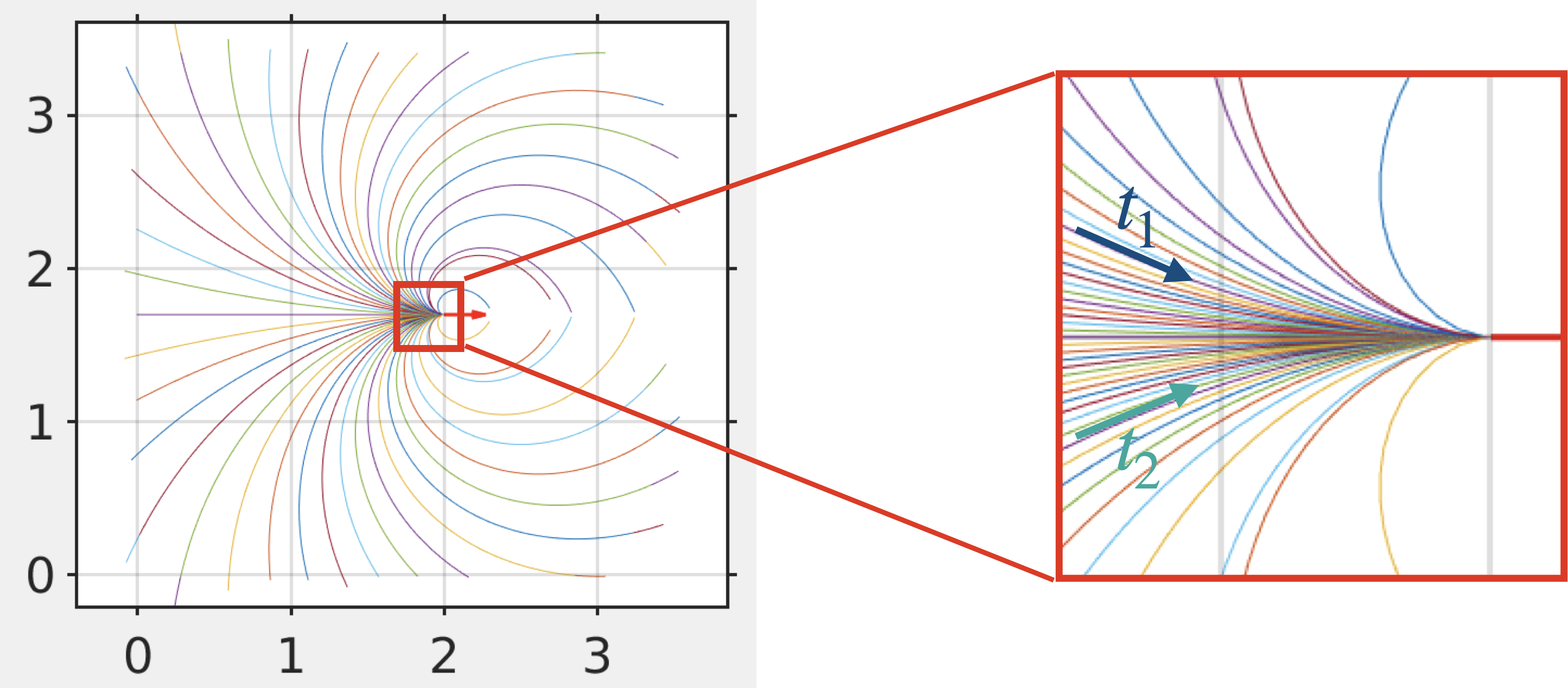}
\caption[]{ The plots show paths in 2D generated by the CLF of \cite{Park2DCLF, ParkRRT} for Dubins cars. At each point, the tangent to a path ($t_1$ and $t_2$ in the blowup) is the heading angle for the robot. These paths clearly fail to account for a biped's ability to move laterally. Moreover, in practice, an underactuated robot such as Cassie Blue would experience chattering in the heading angle when approaching the goal (red arrow). Moreover, if the robot overshoots the goal, it would have to walk along a circle to return to the goal. For these reasons, a new CLF is needed.}
\label{fig:CurrentCLFProblem}%
\end{figure}

\begin{figure}[t]
\centering
\includegraphics[width=0.9\columnwidth]{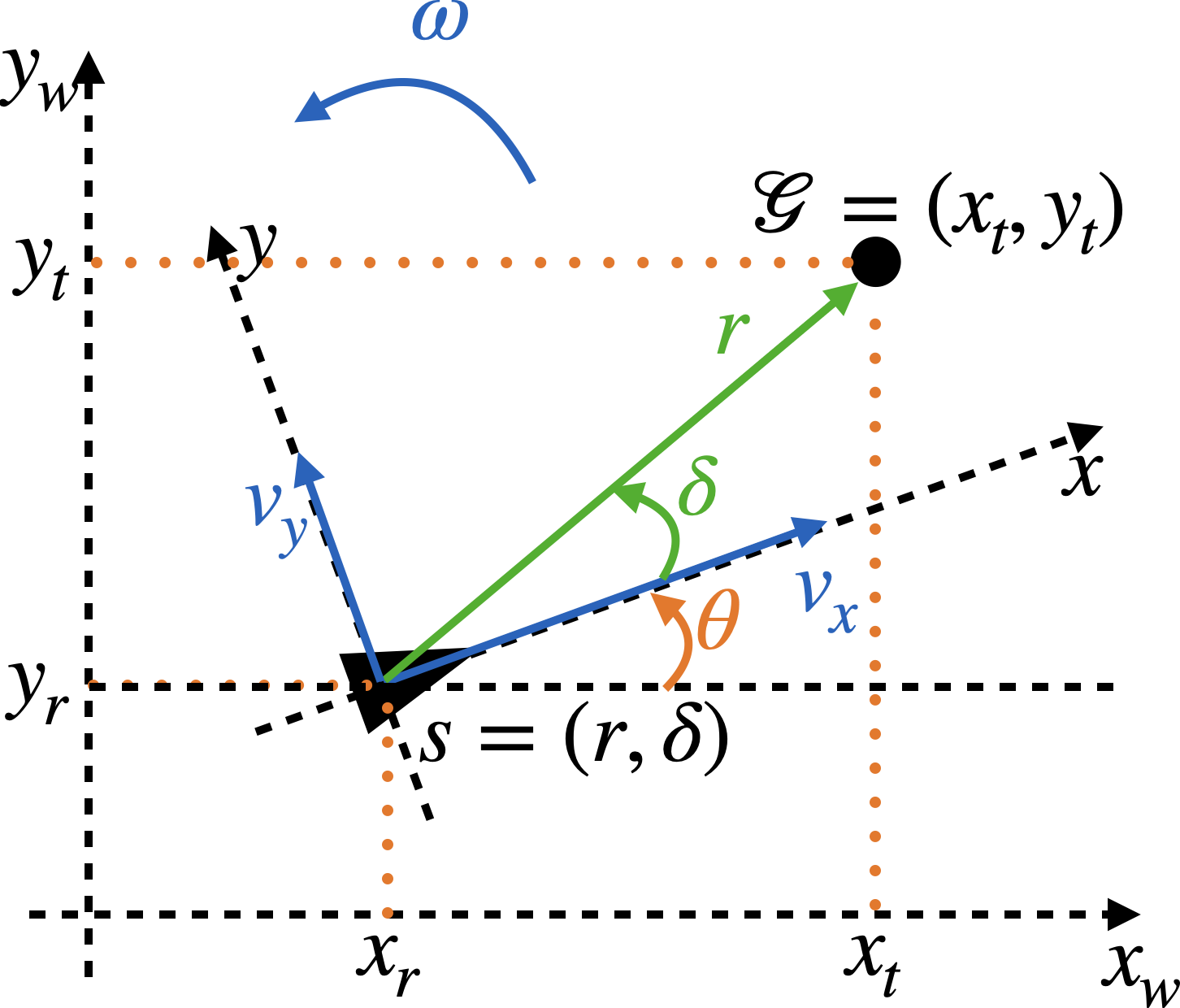}
\caption{Illustration of the robot pose-centric polar representation $(s =
    (r,\delta))$ for robot pose $(x_r, y_r, \theta)$ and target position $\Gcal =
    (x_t, y_t)$. Here, $r$ is the radial distance to the target and $\delta$ is the angle
    between the heading angle $\theta$ of the robot and the line of sight from the
    robot to the goal. The longitudinal velocity, lateral
velocity, and angular velocity are $v_x, v_y$ and $\omega$, respectively.} 
\label{fig:SymbolDefinition}
\end{figure}

\begin{figure}[t]%
\centering
\begin{subfigure}{0.5\columnwidth}
    \centering
\includegraphics[height=0.65\columnwidth]{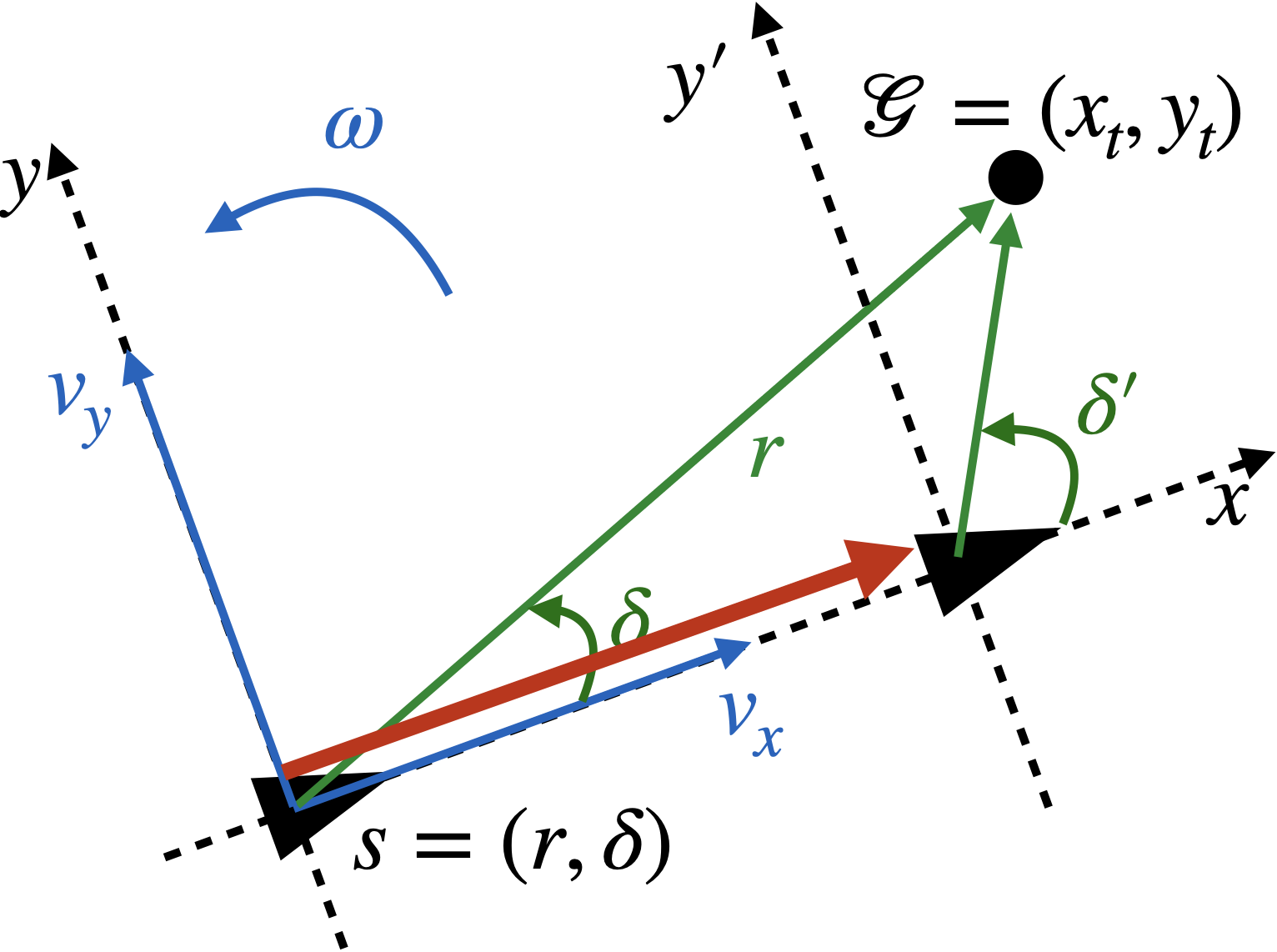}~
\caption{}
\label{fig:DeltaIncreases}%
\end{subfigure}%
\begin{subfigure}{0.5\columnwidth}
    \centering
\includegraphics[height=0.65\columnwidth]{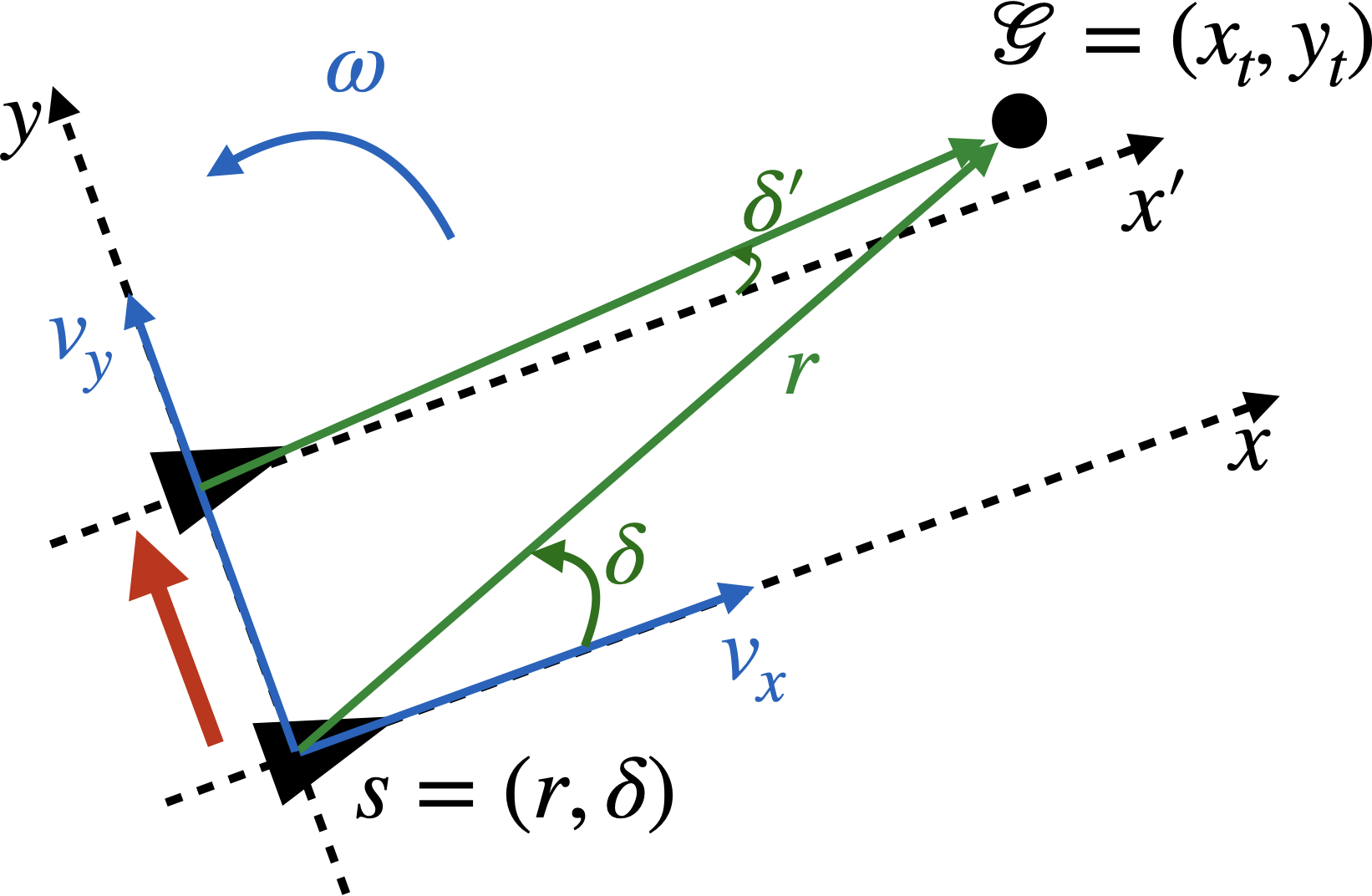}
\caption{}
\label{fig:DeltaDecreases}%
\end{subfigure}%
\caption[]{This figure explains the signs in \eqref{eq:RobotState}.
On the left, $\delta$ increases when the robot moves parallel to the \xaxisN.
Therefore, $\frac{v_x}{r} \sin \delta$ is positive. Similarly, $\frac{v_y}{r} \cos
\delta$ is negated because $\delta$ decreases when the robot moves toward the 
\yaxisN.
}%
\label{fig:DeltaSign}%
\end{figure}

\section{Construction of a Control Lyapunov Function}
\label{sec:CLF}
This section describes the reasons for creating a new CLF function, the construction
of the CLF, and an analysis of its parameters.

\subsection{Deficiency of Existing CLFs}
\label{sec:DeficiencyOfExistingCLF}
The existing 2D CLF planners \cite{ParkRRT, Park2DCLF} are designed for differentially
driven non-holonomically constrained robots, whose dynamics and control laws are inappropriate for bipedal robots. Like most robot
models, this work assumes that the robot is able to continuously change its velocity and heading. However, this is not possible for underactuated bipeds such as Cassie Blue. According to the
ALIP model used for low-level feedback control of Cassie Blue \cite{gong2020angular, 131811, BLICKHAN19891217, 898695, dai2014whole}, the heading angle and the longitudinal and lateral velocity commands can only be updated at the initiation of a step and not within a step. In other words, bipedal robots such as Cassie are not able to
update their control commands during the swing phase and will execute the received
control commands for the entire swing phase. When piece-wise constant commands are applied to the existing 2D CLF of \cite{ParkRRT, Park2DCLF}, built around a Dubins car model, the closed-loop system will oscillate about the discrete heading directions as the robot approaches the goal pose, as explained in Fig.~\ref{fig:CurrentCLFProblem}. This oscillation is undesirable as it can affect the robot's balance.

With a Dubins car model as used in \cite{ParkRRT, Park2DCLF}, the linear velocity is always aligned with the heading angle of the vehicle, and hence this is also true as the vehicle approaches an equilibrium pose. Consequently, a CLF for a target position must also include a target heading, therefore, a target pose. The vehicle must steer and align itself as it approaches the target. Cassie Blue, on the other hand, similar to an omnidirectional robot, is able to move laterally with zero forward velocity, which allows the robot to 
start with an arbitrary pose and arrive at a goal position with an arbitrary heading (i.e., start
with a pose and end with a position). Lateral walking, however, requires more effort due to the limited workspace of the lateral hip joints on the robot and this should be taken into account when designing a CLF. 

To avoid undesirable oscillating movement and account for lateral walking, a new candidate CLF is designed on the basis of an appropriate kinematics model for underactuated bipeds and other omnidirectional robots.

\subsection{State Representation}
As mentioned in Sec.~\ref{sec:CLF}, Cassie Blue is able to walk in any direction. Therefore, we
model Cassie Blue as an omnidirectional robot and reduce it to a directional point
mass. We will account for the increased effort required to walk laterally when we design the CLF.

Denote $\Pcal = (x_r,y_r, \theta)$ the robot pose and $\Gcal = (x_t,y_t)$ the goal
position in the world frame. Let $s$ be the state of an omnidirectional robot
represented in a \textit{robot pose-centric polar coordinate}:
\begin{equation}
    s = \{(r,\delta)| r \in \reals, \text{ and } \delta \in (-\pi, \pi]\},
\end{equation}
where $r = \sqrt{(x_t - x_r)^2 + (y_t - y_r)^2}$ and $\delta$ is the angle
between the heading angle of the robot $(\theta)$ and the line of sight from the
robot to the goal, as shown in Fig.~\ref{fig:SymbolDefinition}.

\begin{remark}
References \cite{ParkRRT, Park2DCLF} used \textit{target pose-centric} polar coordinates because the wheelchair robot needed to arrive at a target position with a target heading angle. In our case, we can use \textit{robot pose-centric} coordinates because we have the freedom to arrive at the target position with any heading angle.
\end{remark}

\subsection{Construction of Control Lyapunov Function }
\label{sec:CLFConstruction}
The kinematics of an omnidirectional robot is defined as
\begin{equation}
\label{eq:RobotState}
    \begin{bmatrix}
    \dot{r} \\
    \dot{\delta}
    \end{bmatrix} = \left[ \begin{array}{rr} -\cos(\delta) &-  \sin(\delta) \\
    \frac{1}{r}\sin(\delta) &  - \frac{1}{r} \cos(\delta) \end{array} \right]
    \begin{bmatrix} v_x \\ v_y
    \end{bmatrix} + \begin{bmatrix}
    0 \\ \omega
    \end{bmatrix}.
\end{equation}
In the above expression, we view $v_x$, $v_y$ and $\omega$ as control variables.
Because the matrix $$ \left[ \begin{array}{rr} -\cos(\delta) &-  \sin(\delta) \\
\frac{1}{r}\sin(\delta) &  - \frac{1}{r} \cos(\delta) \end{array} \right]$$ is
negative definite (and hence invertible) for all $r>0$, the model \eqref{eq:RobotState} is over actuated for
$r>0$.

\begin{remark}
Observe that $\delta < \delta^\prime$ when the robot moves along the \xaxisN, as
shown in Fig.~\ref{fig:DeltaIncreases}. Therefore, $\frac{v_x}{r} \sin \delta$ is
positive. Similarly, $\frac{v_y}{r} \cos \delta$ is negated because $\delta <
\delta^\prime$ when the robot moves toward the \yaxisN, as shown in
Fig.~\ref{fig:DeltaDecreases}.
\end{remark}

We next note that the change of control variables
\begin{equation*}
\label{eq:LinearizingFeedback}
\begin{bmatrix}
v_r\\
v_\delta
\end{bmatrix} :=
  \begin{bmatrix}
\cos \delta & \sin\delta\\
\frac{\sin\delta}{r} & -\frac{\cos \delta}{r}
\end{bmatrix}
\begin{bmatrix}
v_x\\
v_y
\end{bmatrix} + \begin{bmatrix}
    0 \\ \omega
    \end{bmatrix},
\end{equation*}
allows us to feedback linearize the model to a pair of integrators

\begin{equation*}
\label{eq:RobotStateFeedbackLinearized}
    \left[ \begin{array}{r}
    \dot{r} \\
    \dot{\delta}
    \end {array}\right]= \left[ \begin{array}{r} -v_r \\ v_\delta\end{array} \right] .
\end{equation*}
We note that for this model, any positive definite quadratic function is
automatically a CLF. For later use, we note that for all $r>0$,
\begin{equation}
\label{eq:LinearizingFeedbackInverse}
\begin{bmatrix}
v_x\\
v_y
\end{bmatrix} :=
  \begin{bmatrix}
\cos(\delta) & r \sin(\delta)\\
\sin(\delta) & -r \cos(\delta)
\end{bmatrix}
\begin{bmatrix}
v_r\\
v_\delta - \omega
\end{bmatrix}.
\end{equation}

As mentioned in Sec.~\ref{sec:CLF}, lateral walking is more expensive than
longitudinal walking because movement in the lateral hip joint is limited. A candidate control Lyapunov function\footnote{In polar coordinate, the function $\ell$ is positive definite in the sense that $\ell=0 \implies r=0$, and when $r=0$, the angle $\delta$ is arbitrary or undefined.} 
$\ell$, in terms of the robot's current pose and target (end) position, is defined as
\begin{equation}
    \label{eq:CLFDistance}
    \ell= \frac{r^2 + \gamma^2 \sin^2(\beta\delta)}{2},
\end{equation}
where $\gamma$ is a weight on the orientation and the role of $\beta >0$ will be described later. We next check that $\ell$
is a Control-Lyapunov function. The derivative of $\ell$ is
\begin{equation}
    \begin{aligned}
        \dot{\ell}&= r \dot{r} + \frac{\beta\gamma^2}{2} \,\sin\left(2\,\beta
        \,\delta \right)\dot{\delta} \\
                  &= r (-v_r) + \frac{\beta\gamma^2}{2} \sin(2
        \beta\delta) v_\delta.
    \end{aligned}
\end{equation}
The feedback 
\begin{equation}
\label{eq:NewCLFConditions1}
    \begin{aligned}
    v_r &= k_{r1} \frac{r}{k_{r2} +r}\\
    v_\delta &=  -\frac{2}{\beta} k_{\delta 1} \frac{r}{k_{\delta 2} + r}
\sin(2\beta\delta)
\end{aligned}
\end{equation}
results in
\begin{equation}
\label{eq:DNewCLFConditions1}
 \dot{\ell}=  -\frac{k_{r1}}{k_{r2} +r}r^2 - k_{\delta 1} \beta\gamma^2\frac{r}{k_{\delta 2} + r}\sin(2
        \beta\delta),
\end{equation}
which is negative for all $r>0$, $\beta>0$, $k_{r1}>0,~k_{r2}>0,~k_{\delta 1}>0,\text{ and }k_{\delta 2}>0$.

\begin{remark}
\label{rmk:CLFManifoldProperties}
From \eqref{eq:NewCLFConditions1}, it follows that $\dot{\delta}=0$ for $2 \beta \pi \in \{0, \pm \pi\}$. Therefore, the manifolds 
$$M_\delta:=\{ (r, \delta) ~|~ r\ge 0, \delta \in \{ 0, \frac{\pi}{\beta},   \pm \frac{\pi}{2\beta}\}\}$$
are invariant for the closed-loop system.
From \eqref{eq:DNewCLFConditions1}, the manifold $M_\delta$ is locally attractive for $\delta \in \{0, \frac{\pi}{\beta} \}$ and repulsive for $ \delta = \pm \frac{\pi}{2\beta}$. By selecting $\beta>0$, the repulsive invariant manifold can be placed outside the field of view (FoV) of Cassie, as shown in
Fig.~\ref{fig:DifferentStartingPoses}. In practice, a finite-state machine (FSM) is needed so that the robot will initially turn in place so that it starts with the goal located within the FoV
of its sensor suite. 
\end{remark}

The next step is to set up an optimization such that the control
variables $(v_x, v_y, \omega)$ satisfy \eqref{eq:NewCLFConditions1} and take into account that walking sideways takes more effort than walking forward, for Cassie. Because the camera faces forward, walking backward is only selected if the robot is localized into an already built portion of the map.



\subsection{Closed-form Solution}
Taking \eqref{eq:LinearizingFeedbackInverse} as a constraint, we propose to select $\omega$ so as to keep $v_y$ small (limit lateral walking) by optimizing
\begin{equation}
    \label{eq:optimization}
    J = \min_{v_y, \omega}~~ (v_y)^2 + \alpha \omega^2.
\end{equation}
The parameter 
$\alpha>0$ allows us to penalize aggressive yaw motions $\omega$, as will be illustrated in Sec.~\ref{sec:CLFStudy}. Plugging in the constraint \eqref{eq:LinearizingFeedbackInverse}, \eqref{eq:optimization} leads to 
\begin{align*}
    J &= \min_{\omega}~~\{\left[ \sin(\delta) v_r -r \cos(\delta) (v_\delta - \omega)
                          \right]^2 + \alpha \omega^2\}\\ 
      &= \min_{\omega}~~\{\left(\sin(\delta)
                v_r\right)^2  + \left[r \cos(\delta)
(v_\delta - \omega) \right]^2 \\
    &~~~~~~~~~~~- 2 \sin(\delta) v_r (r \cos(\delta) (v_\delta -
\omega)) + \alpha \omega^2\}. 
\end{align*}
A few algebraic calculations and the dropping of ``constant terms'' lead to
\begin{align*}
    \omega^\ast = \argmin_{\omega}~~\{&r^2 \cos^2(\delta) \left(v_\delta - \omega \right)^2 +\\ 
                                      &2 r v_r \sin(\delta) \cos(\delta)\omega + \alpha \omega^2\},
\end{align*}
which implies that
$$\left( \alpha + r^2 \cos^2(\delta) \right) \omega^\ast + r \cos(\delta) \left[ v_r \sin(\delta) - r v_\delta \cos(\delta) \right] = 0$$.
The final result is
\begin{equation}
\label{eq:CLFOmega}
    \omega^* = \frac{ r \cos(\delta) \left[ r v_\delta \cos(\delta) - v_r \sin(\delta) \right]}{\alpha + r^2 \cos^2(\delta) },
\end{equation}
and then
\begin{equation}
\begin{aligned}
    \label{eq:CLFVxVy}
v_y^\ast &= \frac{\alpha \,\left(v_{r}\,\sin\left(\delta \right)-r\,v_{\delta }\,\cos\left(\delta \right)\right)}{r^2\,{\cos\left(\delta \right)}^2+\alpha }\\
v_x^\ast &= \frac{v_{r}\,\cos\left(\delta \right)\,r^2+\alpha \,v_{\delta }\,\sin\left(\delta \right)\,r+\alpha \,v_{r}\,\cos\left(\delta \right)}{r^2\,{\cos\left(\delta \right)}^2+\alpha }.
\end{aligned}
\end{equation}

\begin{figure}[tb]%
\centering
\includegraphics[width=1\columnwidth]{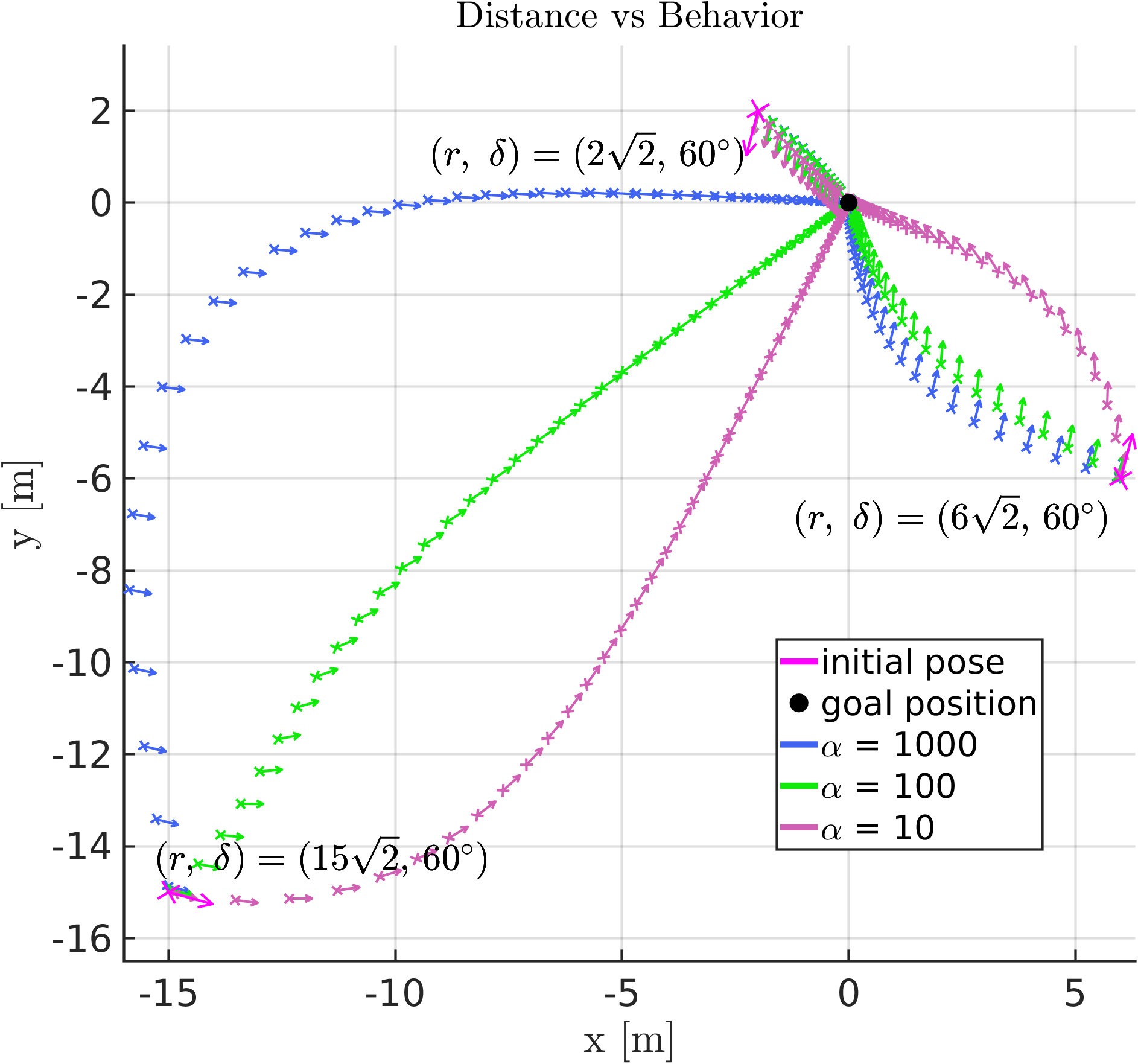}
\caption[]{The distance to the target and the penalty on yaw motion in \eqref{eq:optimization} both affect closed-loop behavior arising from the CLF. The arrows indicate the robot's absolute heading. In each solution of the closed-loop system, the robot's heading relative to the target is initialized at ${60^\circ}$.
When the robot is distant from the goal and
    the heading does not point toward the goal, it will align its relative heading to the target while approaching
     the goal. The level of alignment depends on the yaw motion penalty, $\alpha$. On the other hand, when the robot is close to the goal, the closed-loop controller no longer adjusts the heading angle and employs a lateral motion to reach the goal.
}
\label{fig:DistanceVSBehavior}%
\end{figure}

\begin{table}[t]
\centering
\caption{The default values of parameters.}
\label{tab:ControlParameters}
\scalebox{1}{
\begin{tabular}{|c|c|c|c|c|c|c|}
\hline
$\alpha$ & $\beta$ & $\gamma$ & $k_{r1}$ & $k_{r2}$ & $k_{\delta1}$ & $k_{\delta2}$ \\ \hline
10       & 1.2                    & 1        & 1        & 5        & 0.1           & 10            \\ \hline
\end{tabular}
}
\end{table}

\subsection{Qualitative Analysis of the Closed-loop Trajectories}
\label{sec:CLFStudy}
The default parameters applied in this analysis are shown in Table.~\ref{tab:ControlParameters}. Figure~\ref{fig:DistanceVSBehavior} shows how the closed-loop trajectories vary as a function of heavy, medium, and light penalties on yaw motion, and three different initial distances from the target, with $\delta$, the robot's heading relative to the target, fixed at ${60^\circ}$. We observe that with $r=2 \sqrt{2}$, the robot walks laterally to achieve the goal for all values of the penalty on yaw motion. With $r=15\sqrt{2}$ and $\alpha = 10$, the robot aligns its heading to the target while walking to reduce its lateral movement, whereas with $\alpha = 100$, it maintains its heading and combines lateral and longitudinal motion as needed to reach the goal.  

\begin{figure}[t]%
\centering
\includegraphics[width=1\columnwidth]{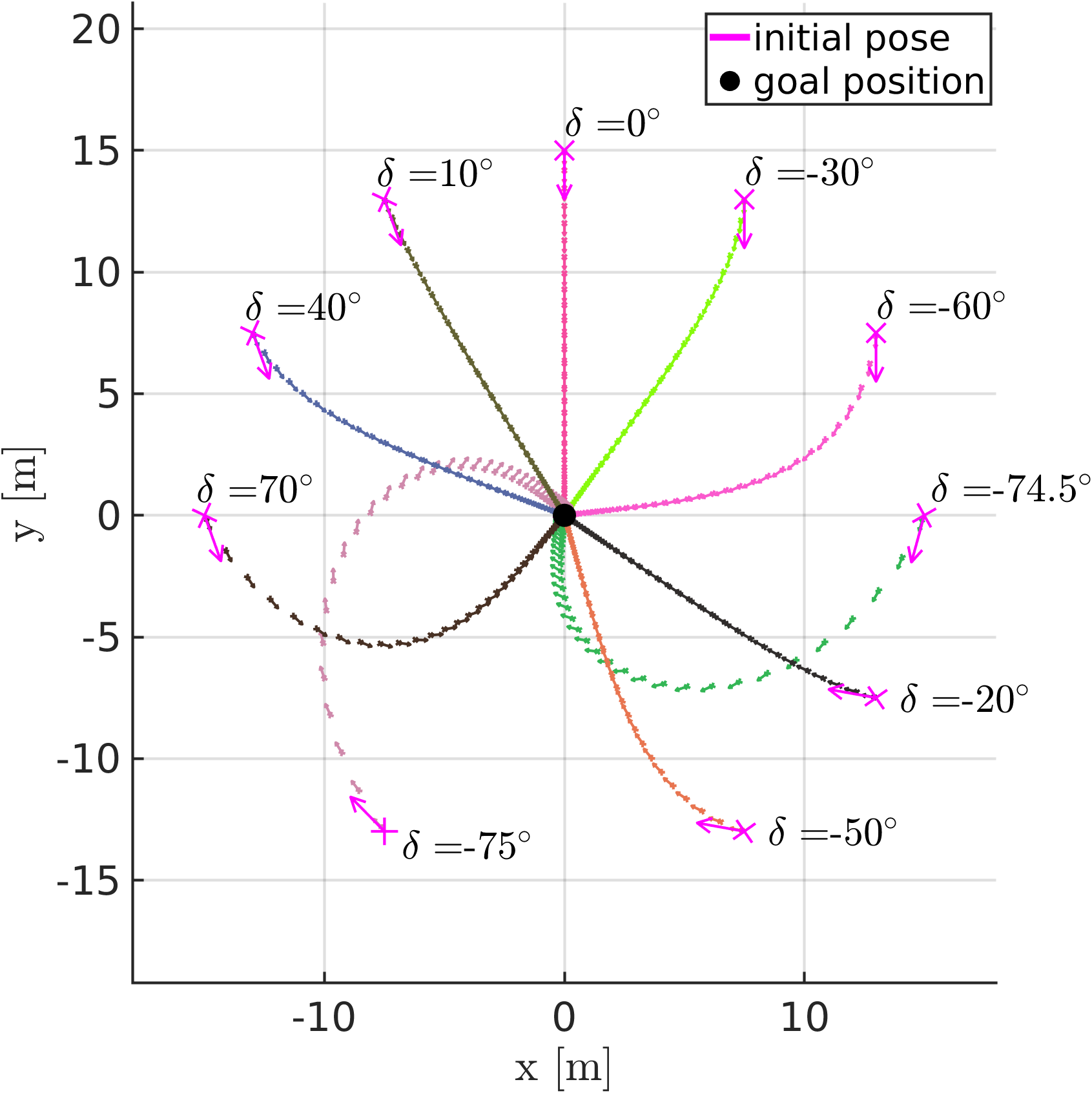}
\caption[]{This figure illustrated how the closed-loop trajectories generated by the CLF in \eqref{eq:CLFDistance} vary as a function the initial relative heading to the target, when starting at a fixed distance of $r=15$ m, and $\alpha = 10$. The arrows indicate the robot's heading. As shown in Table~\ref{tab:ControlParameters}, we are using $\beta=1.2$, which yields an FoV of $\pm75^\circ$. For relative heading ``errors'' less than $40^\circ$, the robot aligns quickly to the target and longitudinal walking dominates. These motions should be compared to those in Fig.~\ref{fig:DistanceVSBehavior}}
\label{fig:DifferentStartingPoses}%
\end{figure}

Figure~\ref{fig:DifferentStartingPoses} shows how the closed-loop trajectories vary as a function the initial relative heading to the target, when starting at a fixed distance of $r=15$ m, and $\alpha = 10$. As indicated in Table~\ref{tab:ControlParameters}, we are using $\beta=1.2$, which yields FoV of $\pm75^\circ$. For relative heading ``errors'' less than $40^\circ$, the robot aligns quickly to the target and longitudinal walking dominates. If quicker zeroing of the heading error is desired, a smaller value of $\alpha$ could be used or the robot could turn in place before starting a new segment. 

\section{Omnidirectional CLF-\rrtN}
\label{Sec:Planner}
This section integrates the CLF proposed in
Sec.~\ref{sec:CLF} into the original \rrt algorithm. The resulting omnidirectional CLF \rrt provides feasible paths for \eqref{eq:RobotState} while (i) accounting for relative heading, (ii) the asymmetry in roles of target position and current pose induced by the CLF, and (iii) the fact that walking laterally is more challenging than walking in the longitudinal direction for robots such as Cassie.

\subsection{Standard \rrt Algorithm}
The original \rrt\cite{RRTStar} is a sampling-base, incremental planner with
guaranteed asymptotic optimality. In configuration space, \rrt grows a tree where leaves are
states connected by edges of linear path segments with the minimal cost. Additionally, \rrt considers nearby nodes of a sample to choose the best parent node and to rewire the graph if shorter path is possible to guarantee asymptotic optimality.

\subsection{Omnidirectional CLF-\rrt Algorithm}
The omnidirectional CLF-\rrt differs from the original \rrt in four aspects. First, the
distance between two nodes is defined by the CLF in \eqref{eq:CLFDistance}, which takes relative heading into account. Second, the steering/extending
functions use the closed-loop trajectories generated by \eqref{eq:CLFVxVy} to define paths between nodes. Third, because the cost \eqref{eq:CLFDistance} between two nodes $i$ and $j$ is not symmetric (i.e, a different cost is assigned if node $i$ is the origin versus it is the target), a distinction must be made between near-to nodes and near-from nodes. The above three aspects are common to the CLF-\rrt variant introduced in \cite{Park2DCLF, ParkRRT}. Finally, when connecting, exploring, and rewiring the tree, an additional term is added to the cost \eqref{eq:CLFDistance} to account for the relative ease or difficulty of traversing the path.

Our proposed \rrt modification is summarized below with notation that generally
follows~\cite{RRTStarICRA}. Let $\Xcal = \{(x,y,\theta)~|~ x,y\in\reals \text{ and }
\theta\in(-\pi, \pi]\}$ be the configuration space and let $\Xcal_{\text{obs}}$ denote
the obstacle region, which together define the free region for walking $\Xcal_{\text{free}} = \Xcal\backslash
\Xcal_{\text{obs}}$. The omnidirectional CLF \rrt solves the optimal path planning
problem by growing a tree $\Tcal = (V, E)$, where $V\in\xfree$ is a vertex set of
poses connected by edges $E$ of feasible path segments. Briefly
speaking, the proposed \rrt (Algorithm~\ref{alg:rrt}) explores the configuration
space by random sampling and extending nodes to grow the tree (explore the
configuration space), just as in the classic RRT\cite{VanillaRRT}. Considering nearby nodes of
a sample to choose the best parent node and rewiring the graph guarantee asymptotic
optimality (Algorithm~ \ref{alg:rrtParent} and \ref{alg:rewire}), as with the classic algorithm. As emphasized previously, a key difference lies in how the paths between vertices are generated.

\subsubsection{Sampling} 
This step randomly samples a pose $n_{\text{rand}} =
(x,y,\theta)\in\xfree$. To facilitate faster convergence and 
to find better paths, we use sampling with a goal bias, limited search space, and
Gaussian sampling. For more details, see our implementation on
GitHub~\cite{githubFileCLFPlanning}.

\subsubsection{Distance} 
The distance $d(n_i, n_k)$ from node $n_i$ to node $n_k$ in the tree $\Tcal$ is
defined by \eqref{eq:CLFDistance}, which takes relative heading into account. Note that when computing the distance, $n_i$ is a pose $(x_i, y_i, \theta_i)$ and the heading of $n_k$ is ignored and only its $(x_k, y_k)$ values are used.

\begin{remark}
    As mentioned in Sec.~\ref{sec:CLFConstruction}, the robot will rotate in place if
    the target point is outside the FoV. If rotating in place is laborious, one can
    also consider the following distance function:
\begin{equation}
    d(n_i, n_k) = \ell + k_\delta \max{(|\delta| - |U|, 0)},
\end{equation}
where $\ell$ is defined in \eqref{eq:CLFDistance}, $k_\delta$ is a positive constant,
and $U$ corresponds to a repulsive point (i.e., $\pm \frac{\pi}{2\beta}$) in Remark~\ref{rmk:CLFManifoldProperties}.
\end{remark}

\subsubsection{Traversability of a path}
Let $\Pcal = (x_r, y_r, z_r, \theta)$ be the current robot pose and denote 
$\mathscr{T}(\Pcal, n_i, n_j)$ the path\footnote{The path is
generated from the CLF in Sec.~\ref{sec:CLF}.} connecting $n_i$ and $n_j$.
We also denote $C_e(x_t, y_t), C_s(x_t, y_t)$ as the elevation and the magnitude of the gradient at a point $(x_t, y_t)$ in a local map $\Mcal$, see
Sec.~\ref{sec:PlanningThread}. Finally, let $\mathbb{T}(\Pcal, \traj)$ be the cost of
the path traversability, defined as
\begin{equation}
    \begin{aligned}
    \mathbb{T}(\Pcal, \traj) = \sum_{\forall x_t, y_t \in \traj}
    &C_e(x_t, y_t) + k_s C_s(x_t, y_t) + \\ 
    &k_r (C_e(x_t, y_t) - z_r), 
    \end{aligned}
\end{equation}
where $k_s$ and $k_r$ are the corresponding positive coefficients. 

\begin{remark}
    Traversability varies among different types of robots. Additional elements
    can be added as needed to account for different aspects for traversability
    computation. 
\end{remark}

\subsubsection{Cost between Nodes}
Let $c(n_i, n_k)$ be the cost from $n_i$ to $n_k$ in the tree $\Tcal$, defined
as
\begin{equation}
    c(n_i, n_k) =  d(n_i, n_k) + k_t \mathbb{T}(\Pcal, \traj),
\end{equation}
where $k_t$ is the weight of traversability.

\begin{algorithm}[t]
    \caption{$\Tcal = (V,E) \leftarrow$ Omnidirectional CLF \rrt}
\label{alg:rrt}
    $\Tcal \leftarrow$ InitializeTree();

    $\Tcal \leftarrow$ InsertNode($\emptyset, \nsub{init}, \Tcal$);

\For{i=1 \KwTo $N$}
{
    $\nsub{rand}\leftarrow$ Sample($i$)

    $\nsub{nearest}\leftarrow$ Nearest($\Tcal, \nsub{rand}$)

    $(\nsub{new}, \traj^\prime) \leftarrow$ Extend($\nsub{nearest}, \nsub{rand},
    \kappa$)

    \If{ObstacleFree($\traj^\prime$)}
        {
            $\Ncal_T \leftarrow$ NearTo($\Tcal, \nsub{new}, |V|$)

            $\nsub{min} \leftarrow$ ChooseParent($\Ncal_T, \nsub{nearest}, \nsub{new}$)

            $\Tcal \leftarrow$ InsertNode($\nsub{min}, \nsub{new}, \Tcal$)

            $\Ncal_F\leftarrow$ NearFrom($\Tcal, \nsub{new}, |V|$)
            
            $\Tcal\leftarrow$ ReWire($\Tcal, \Ncal_F, \nsub{min}, \nsub{new}$)

        }
}
\Return $\Tcal$
\end{algorithm}

\subsubsection{Nearby Nodes}
Due to the use of the CLF function, the distinction between near-to nodes $\Ncal_T$
and near-from nodes $\Ncal_F$ is necessary.
\begin{equation}
    \begin{aligned}
        \Ncal_T(n_i, \Tcal, \Mcal, m) := \{&n \in V~ | ~d(n, n_i) \le L(m) ~~\& \\
                                      &|\mathbb{T}(n, \Pcal) - \mathbb{T}(n_i, \Pcal)| 
                                  \le T_k\},
    \end{aligned}
\end{equation}
where $|\cdot|$ is the absolute value, $m$ is the number of nodes in the tree $\Tcal$,
and $L(m) = \eta\left(\log(n)/n\right)^{(1/\xi)}$ with the constant $\eta$ and
dimension of space $\xi$ ($3$ in our case)~\cite{karaman2010optimal} and $T_k$ is a
positive constant. Similarly, the near-from nodes $\Ncal_F$ are determined by 
\begin{equation}
    \begin{aligned}
        \Ncal_F(n_i, \Tcal, \Mcal, m) := \{&n \in V ~| ~d(n_i, n) \le L(m) ~~\& \\
                                      &|\mathbb{T}(n, \Pcal) - \mathbb{T}(n_i, \Pcal)|
                                  \le T_k\}.
    \end{aligned}
\end{equation}

\subsubsection{Nearest Node}
Given a node $n_i\in \Xcal$, the tree $\Tcal$, and the local map $\Mcal$, the nearest
node is any node $n_* \in \Tcal$ in the tree where the cost from $n_*$ to $n_i$ is minimum.

\subsubsection{Steering and Extending}
The steering function generates a path segment $\traj$ that starts from $n_i$ and
ends exactly at $n_k$. The extending function extends the path from $n_i$
toward $n_k$ until $n_k$ is reached or the distance traveled is $\kappa$ in which case it returns
a new sample $\nsub{new}$ at the end of the extension.

\begin{algorithm}[t]
    \caption{\scalebox{0.95}{$\nsub{parent}\leftarrow$ ChooseParent($\Ncal_T, \nsub{nearest}, \nsub{new}$)}}
\label{alg:rrtParent}
    $\nsub{parent}\leftarrow\nsub{nearest}$

    $\csub{parent}\leftarrow$ Cost($\nsub{nearest}$)  $+$ c$(\nsub{nearest}, \nsub{new})$

\For{$n_{\text{near}} \in \Ncal_T$}
{
        $\traj^\prime\leftarrow$Steer$(\nsub{near}, \nsub{new})$

        \If{ObstacleFree($\traj^\prime$)}
        {
            $c^\prime = \text{Cost}(\nsub{near}) + c(\nsub{near}, \nsub{new})$

            \If{$c^\prime < \text{Cost}(\nsub{new})$ and $c^\prime < \csub{parent}$}
            {
                $\nsub{parent} \leftarrow \nsub{near}$

                $\csub{parent} \leftarrow c^\prime$
            }
        }
}
\Return $\nsub{parent}$
\end{algorithm}

\subsubsection{Parent Choosing and Graph Rewiring}
Choosing the best parent node (Algorithm~\ref{alg:rrtParent}) and rewiring the graph (Algorithm~\ref{alg:rewire}) guarantee asymptotic optimality.
Let Cost$(n_i)$ be the cost from the root of the tree $\Tcal$ to the node $n_i$. The parent $\nsub{parent}$ of a node $\nsub{new}$ is determined by finding a node $n_i \in \Ncal_T$ with smallest cost from the root to the node:
\begin{equation}
    \nsub{parent}= \argmin_{\nsub{near}\in \Ncal_T} \text{Cost}(\nsub{near}) + c(n_\text{near}, \nsub{new}).
\end{equation}
After a parent node is chosen, nearby nodes $\Ncal_F$ are rewired if shorter paths are found.
In our experiments, we
used the extending function for exploration, and the steering function to find
the best parent node and to rewire the graph.

\subsubsection{Collision Check}
This step verifies whether a path $\traj$ lies within the obstacle-free region of
the configuration space. Note that additional constraints, such as curvature bounds and
minimum clearance, can also be examined in this step. 

\subsubsection{Node Insertion}
Given the current tree $\Tcal = (V, E)$ and a node $v\in V$, this step inserts the 
node $\nsub{}$ to $V$ and creates an edge $e_{nv}$ from $\nsub{}$ to $v$.

\begin{algorithm}[b]
    \caption{$\Tcal \leftarrow$ 
             ReWire$(\Tcal, \Ncal_F,n_{\text{min}}, n_{\text{new}})$}
\label{alg:rewire}
\For{$n_{\text{near}} \in \Ncal_F\backslash\{n_{\text{min}}\}$}
{
        $\traj^\prime\leftarrow$Steer$(\nsub{new}, \nsub{near})$

        \If{ObstacleFree($\traj^\prime$) and \\
        ~~~Cost($\nsub{new}$) $+$ c($\nsub{new}, \nsub{near}$) $<$ Cost($\nsub{near}$)}
        {
            $\Tcal \leftarrow$ Re-Connect($\nsub{new}, \nsub{near}, \Tcal$)
        }
}
    \Return $\Tcal$
\end{algorithm}

\section{Reactive Planning System}
\label{sec:PlannerSystem}

The previous section provides a sparse set of paths from a robot's initial location to a goal. The degree of optimality depends on how long the planning algorithm is run. A typical update rate may be 5 Hz for real-time applications. When the robot is perturbed off the nominal path, one is left with deciding how to reach the goal, say by tracking the nominal path with a PID controller. Important alternatives to this, called a high-frequency reactive planner or a feedback motion planner, were introduced in \cite{golbol2018rg, arslan2019sensor, paternain2017navigation, ExactRobot2016, koditschek1990robot, rimon1990exact, borenstein1989real, koditschek1987exact, gomez2013planning, tedrake2010lqr}. A version based on the work of \cite{Park2DCLF, ParkRRT} will be incorporated into our overall planning system. In addition, we take into account features in a local map.

\subsection{Elements of the Overall Planning System}

The overall objective of the planner system is to replace the commonly used waypoint-following or path-tracking strategies with a family of closed-loop feedback control laws that steer the robot along a sequence of collision-free sub-goals leading to the final goal. In simple terms, as in \cite{tedrake2010lqr,Park2DCLF, ParkRRT}, we populate the configuration space with a discrete set of feedback control laws that steer the robot from local chart about a sub-goal to the sub-goal itself. The collision free property is handled by the low-frequency planner at the current time. Others have used CBFs for this purpose \cite{7782377, 7040372, 7864310, 7798370, xu2015robustness, nguyen2020dynamic}. A finite-state machine (FSM) is integrated into the low-frequency planner to
handle high-level mission requirements such as turning left at every intersection. 

The planner assumes the initial robot pose, a final goal, and real-time map building are provided. It is assumed that the initial robot pose and final goal are initialized in an otherwise featureless metric map, with the robot's initial pose as the origin. The featureless map is filled in by the real-time mapping package \cite{Fankhauser2014RobotCentricElevationMapping,
Fankhauser2018ProbabilisticTerrainMapping, Lu2020BKI} based on collected \lidar and/or camera data.

\begin{figure}[t]
\centering
\includegraphics[width=1\columnwidth]{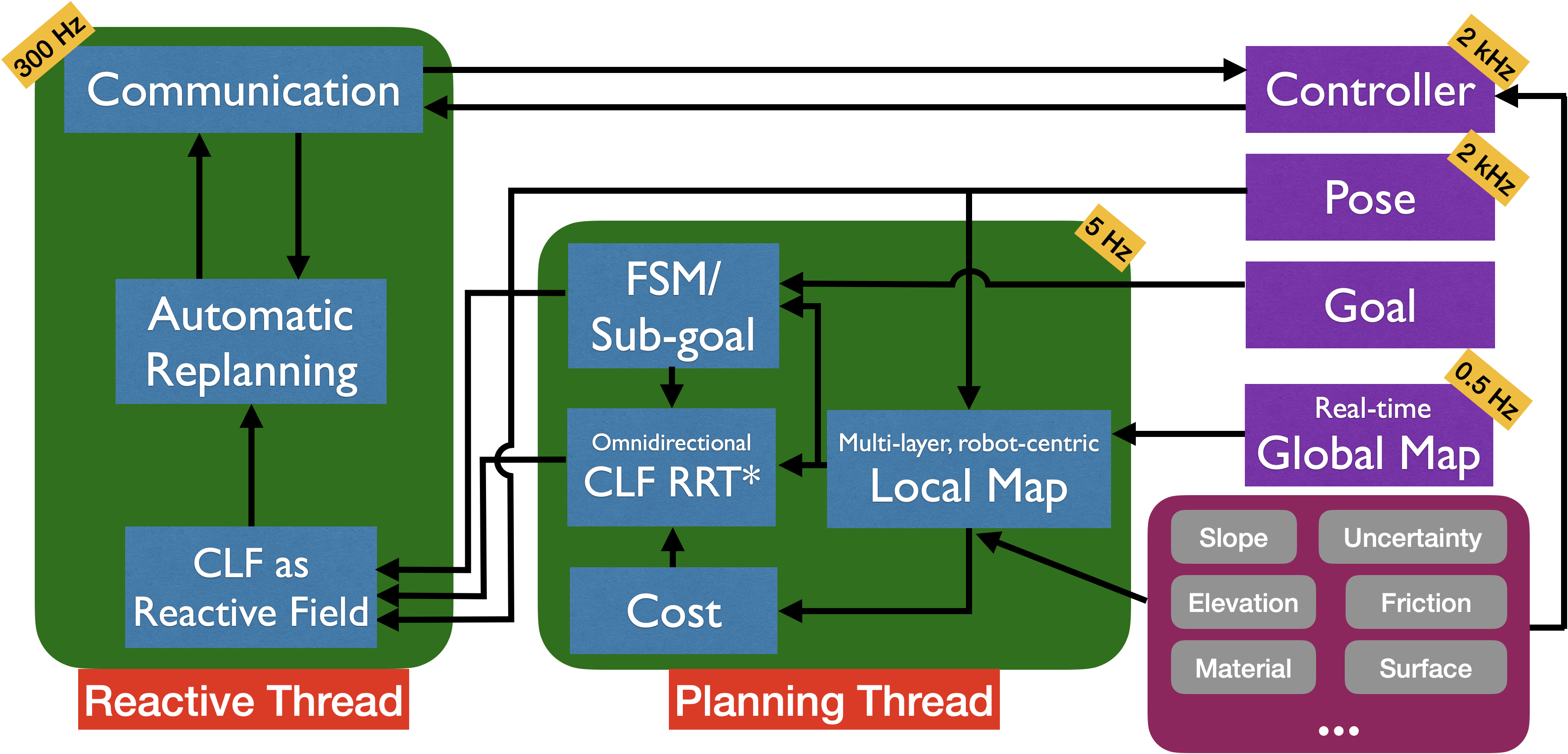}
\caption{This figure summarizes the proposed reactive planning system. The planning thread is built around \rrt and an omnidirectional CLF that is used to assign distances, define locally optimal path segments, search radius, and linking conditions for re-wiring and choosing a parent. In addition, the planning thread contains a
    multi-layer, robot-centric local map for computing traversability, a sub-goal finder, and a finite-state machine (FSM) to choose sub-goal locations guiding the robot to a distant goal. The
    terrain information extracted from the multi-layer local map can be shared with a
terrain-aware controller, such as \cite{gibson2021terrain}.
    Instead of a
    common waypoint-following or path-tracking strategy, the reactive thread copes with robot deviation while eliminating non-smooth motions via a vector field (defined by a
    closed-loop feedback policy arising from the CLF). The vector field provides real-time control commands to the robot's gait controller as a function of instantaneous robot pose.
} 
\label{fig:PlannerSystemDiag}
\end{figure}

\subsection{Planning Thread}
\label{sec:PlanningThread}
The planning thread deals with short-range planning (less than 20 meters) at a
frequency of 5 to 10 Hz. It includes a robot-centric local map, our
omnidirectional CLF-\rrt algorithm of Section~\ref{sec:CLF}, cost computation, a sub-goal finder, and a finite-state machine.

\subsubsection{Robot-centric Local Map and Cost Computation}
\begin{figure}[t]%
\centering
\begin{subfigure}{0.5\columnwidth}
    \centering
\includegraphics[height=0.9\columnwidth]{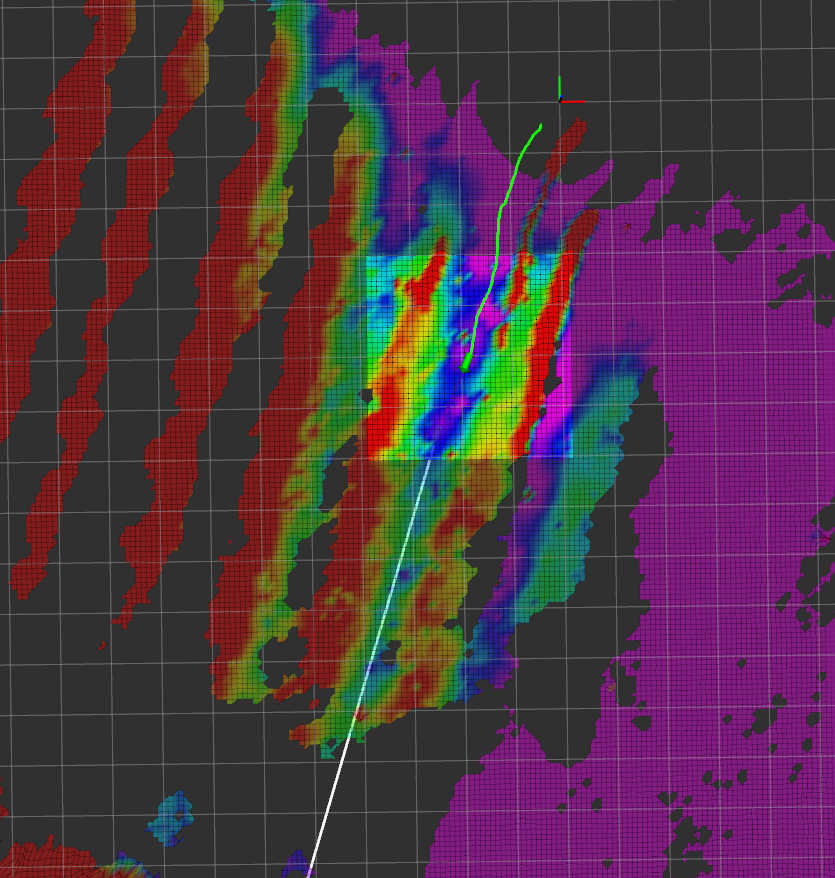}
\caption{}
    \label{fig:LocalMap}%
\end{subfigure}%
\begin{subfigure}{0.5\columnwidth}
    \centering
\includegraphics[height=0.9\columnwidth]{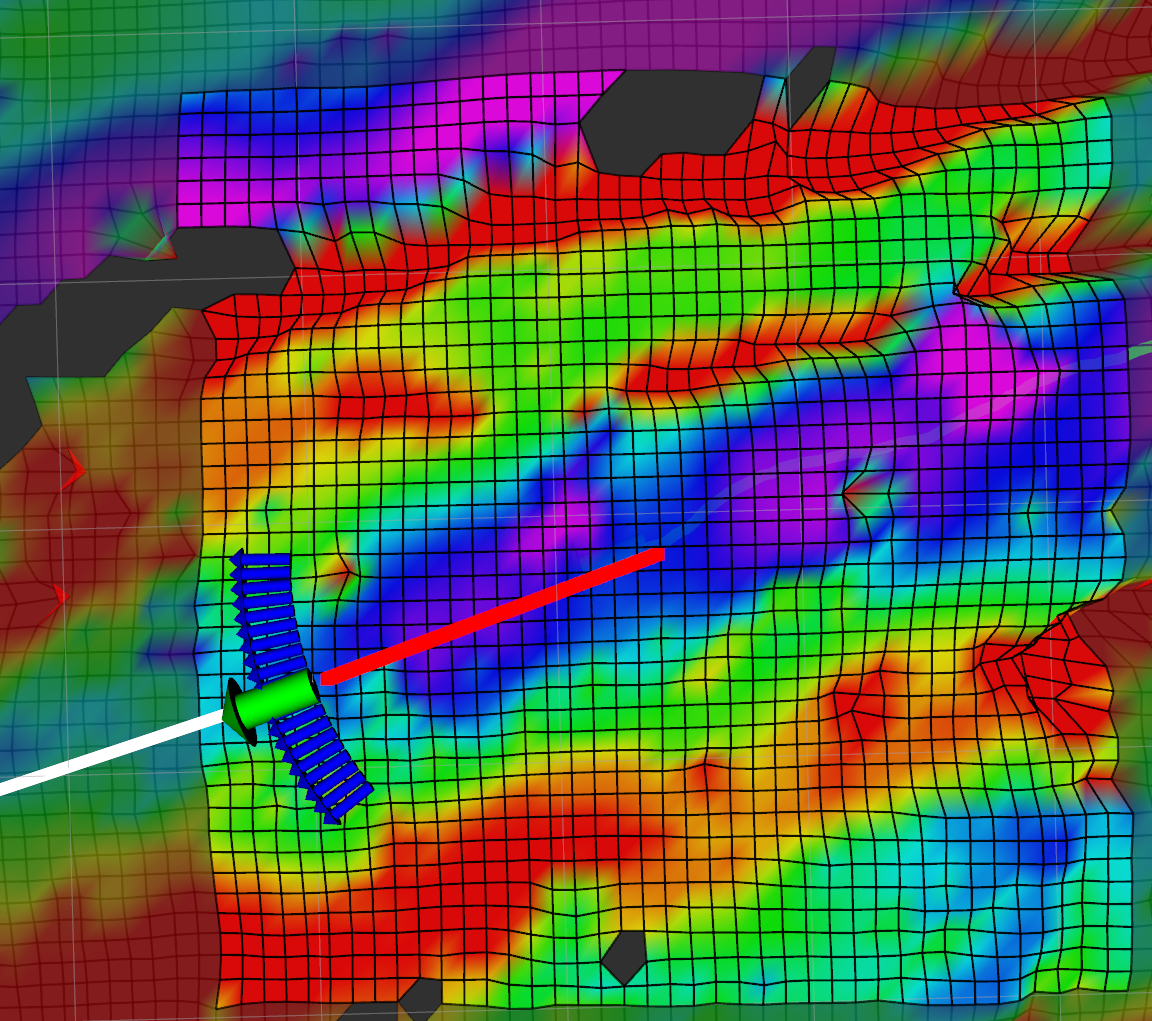}%
\caption{}
\label{fig:ZoomedLocalMap}
\end{subfigure}%
\caption[]{The elevation map was
built online while Cassie was autonomously traversing the Wave Field on the North Campus of the University of Michigan. The highlighted area is the smoothed, robot-centric local map. The blue
    arc and the green arrow are the sub-goal finder and the chosen sub-goal for the
    omnidirectional CLF \rrtN, respectively. The red shows the locally optimal path. }%
\label{fig:LocalMapIllustration}%
\end{figure}

Figure~\ref{fig:LocalMap} shows the robot-centric multi-layer local map (highlighted
area), which crops a sub-map centered around the robot's current position from the
global map provided by the mapping algorithm. The local map computes additional
useful information such as terrain slope (local gradient) which is useful for
assigning cost. Moreover, other necessary operations for different
experiment scenes such as applying the Bresenham algorithm
\cite{bresenham1965algorithm} to remove walkable area behind glass walls can be
computed in this step, see Sec.~\ref{sec:FullyAutonomyExps}.  Additionally,
terrain information such as slopes and frictions can be sent to a terrain-aware low-level
controller \cite{gibson2021terrain}. The computations with
the local map are efficient compared to processing the full map.

\subsubsection{Anytime Omnidirectional CLF-\rrt Planner}
The anytime feature is a direct result of using \rrt as a planner. The algorithm can be queried at anytime to provide a suboptimal path comprised of wayposes, which the CLF \eqref{eq:CLFDistance} turns into real-time feedback laws for anytime replanning.


\begin{figure*}[!t]%
\centering
\begin{subfigure}{0.68\columnwidth}
    \centering
    \label{fig:obsmap1}%
\includegraphics[width=0.97\columnwidth]{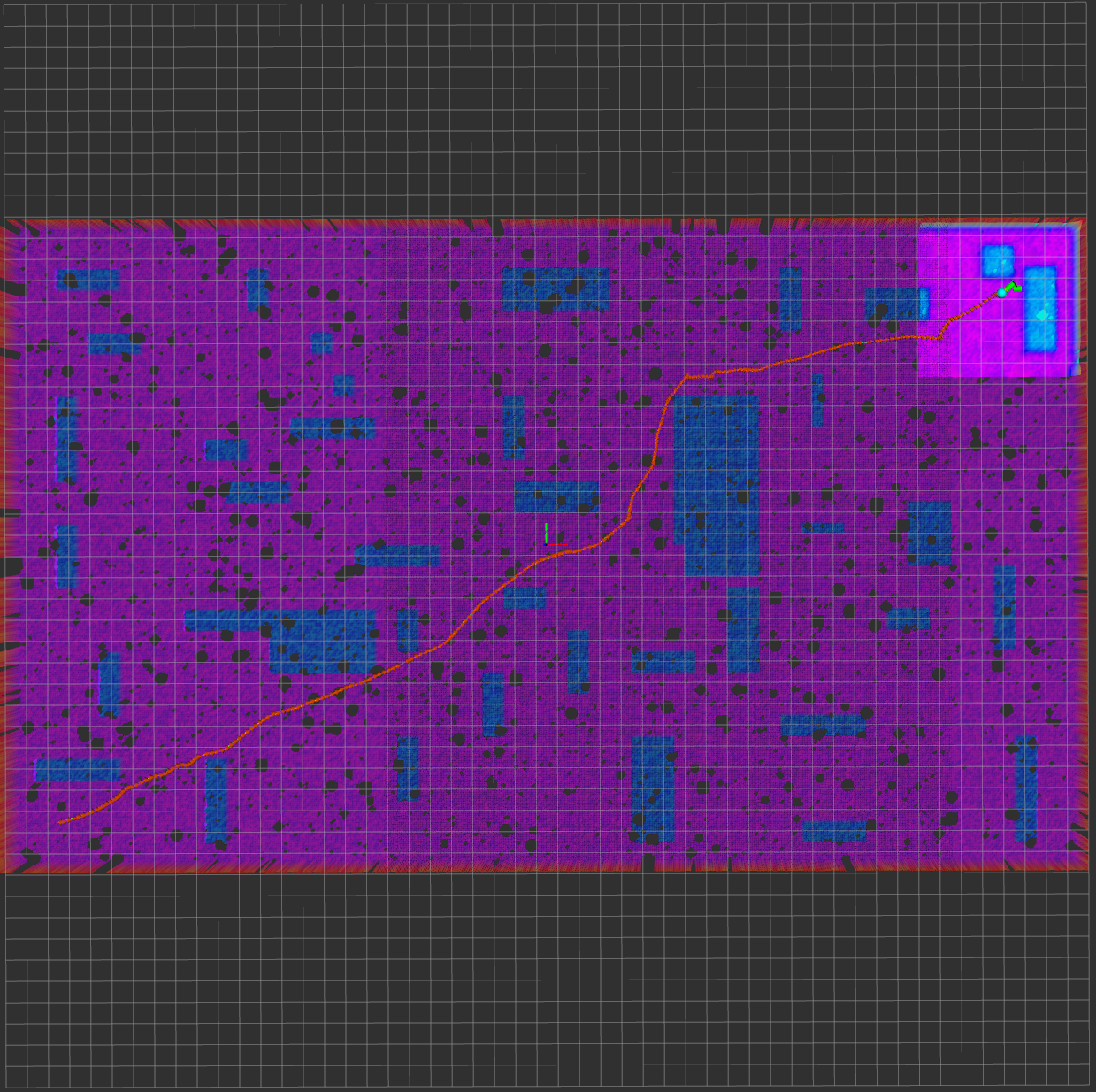}~%
\caption{}
\end{subfigure}%
\begin{subfigure}{0.68\columnwidth}
    \centering
    \label{fig:turnleft1}
\includegraphics[width=0.97\columnwidth]{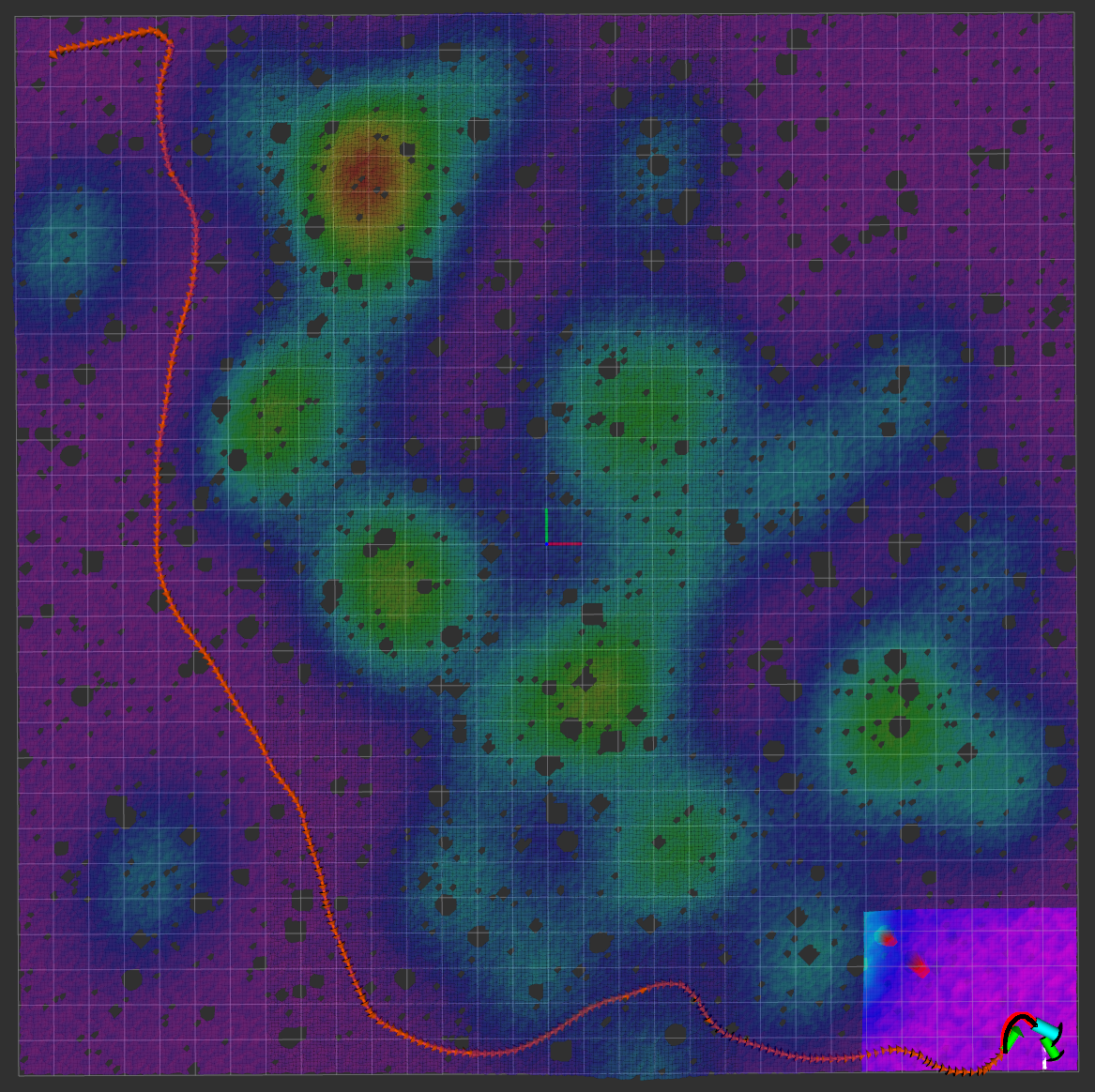}~%
\caption{}
\end{subfigure}%
\begin{subfigure}{0.68\columnwidth}
    \centering
    \label{fig:map1}%
\includegraphics[trim={0 0 0 20},clip, width=0.97\columnwidth]{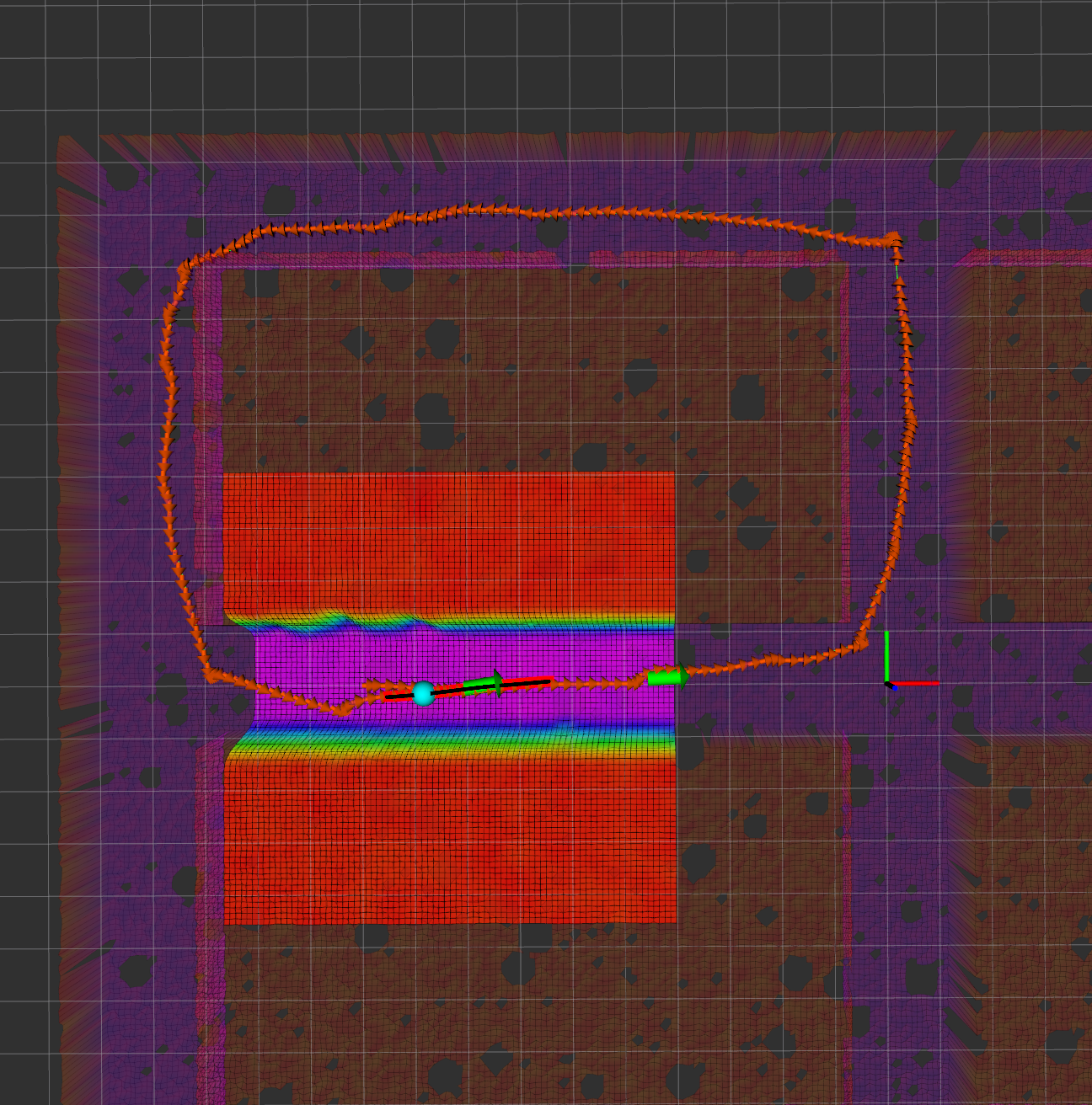}%
\caption{}
\end{subfigure}
\begin{subfigure}{0.68\columnwidth}
    \centering
    \label{fig:obsmap2}%
\includegraphics[width=0.97\columnwidth]{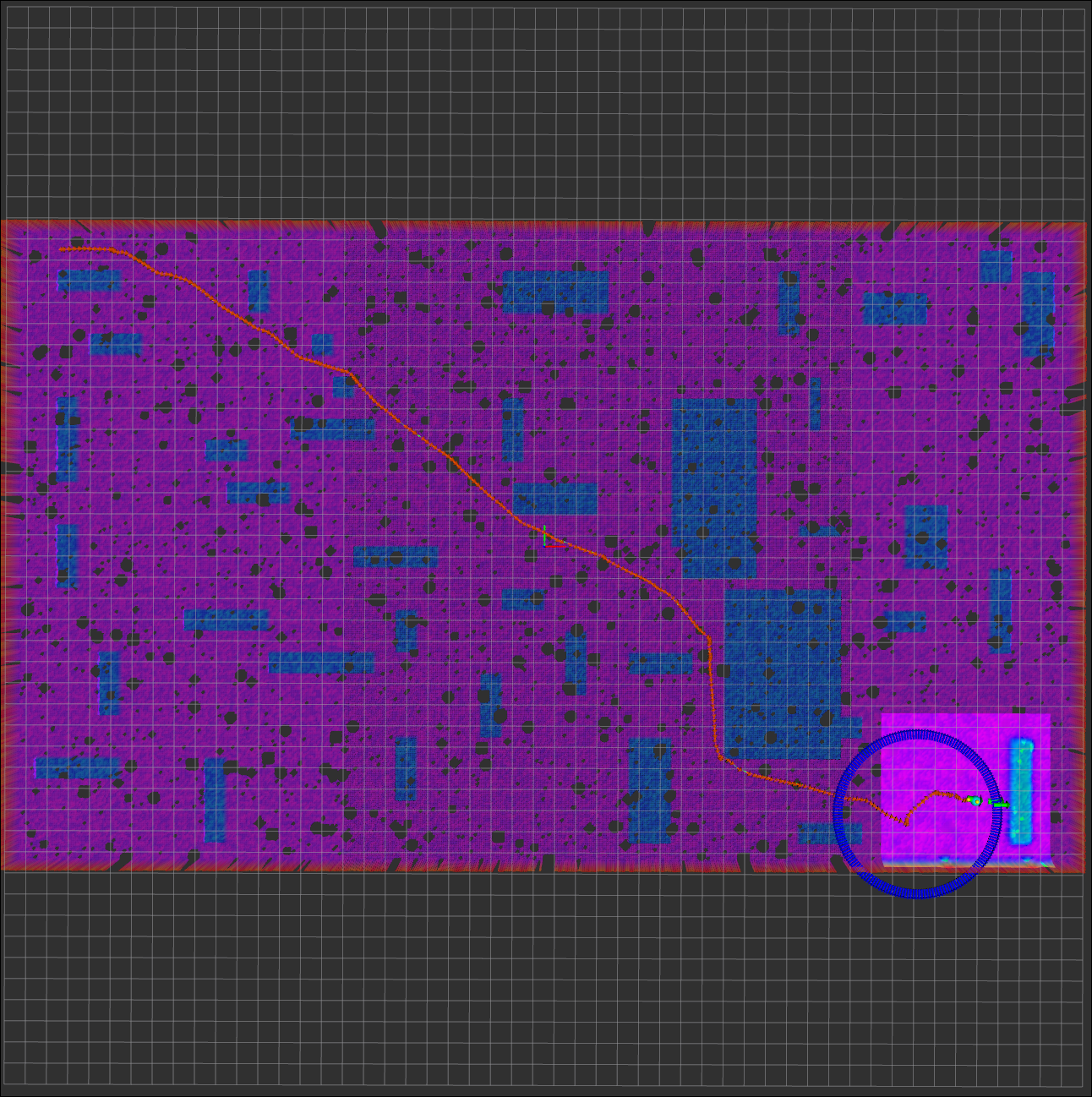}~%
\caption{}
\end{subfigure}%
\begin{subfigure}{0.68\columnwidth}
    \centering
    \label{fig:turnleft2}
\includegraphics[width=0.97\columnwidth]{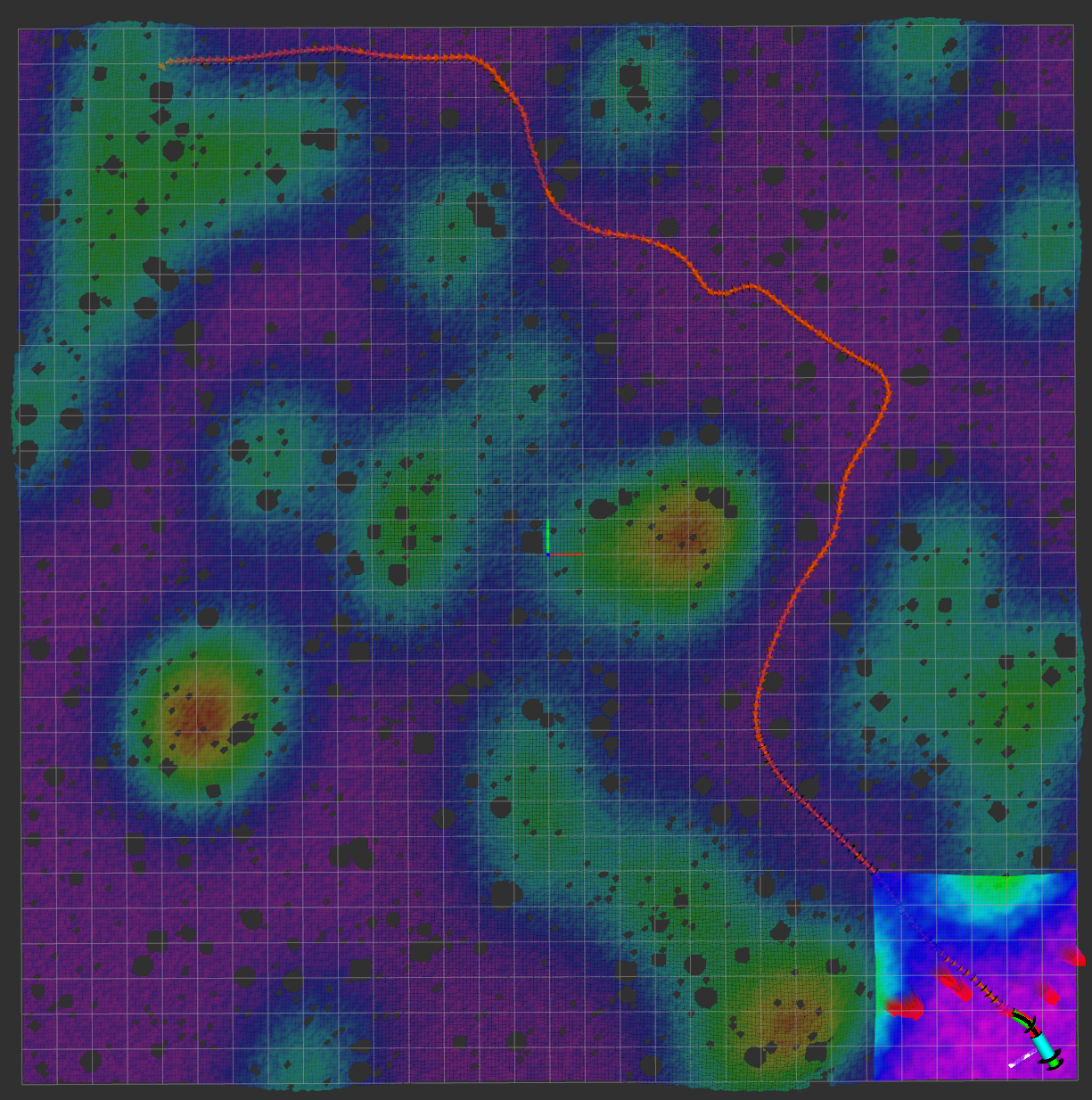}~%
\caption{}
\end{subfigure}%
\begin{subfigure}{0.68\columnwidth}
    \centering
    \label{fig:map3}%
\includegraphics[width=0.97\columnwidth]{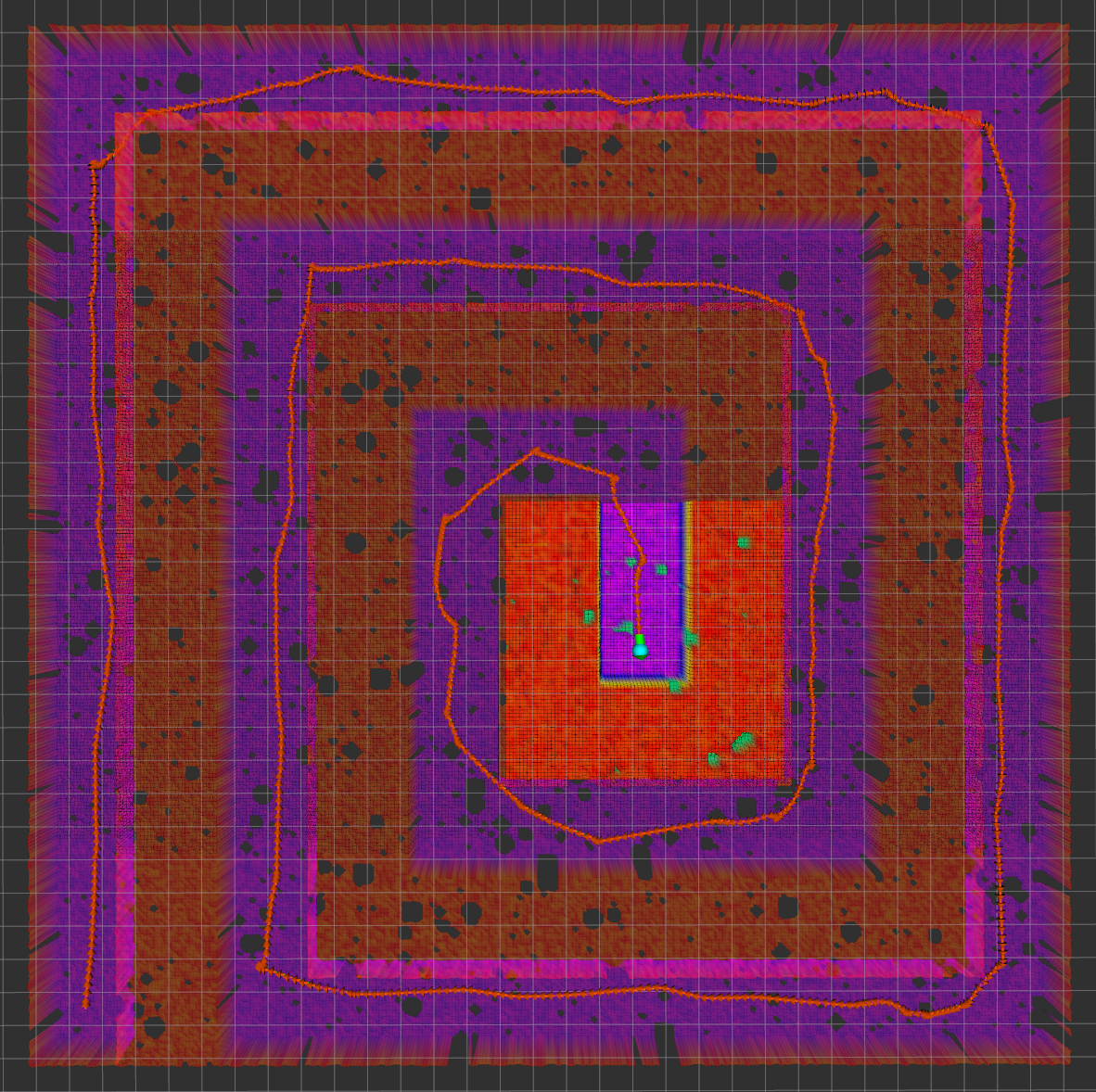}%
\caption{}
\end{subfigure}%
\caption[]{Simulated scenes and results obtained with the proposed reactive planning system. On the left are cluttered indoor scenes with obstacles and holes, in the middle are noisy undulating outdoor terrains, and on the right, are high-level missions. Each grid in a map is $1 \times 1$ meter. The simulated robot is based on the ALIP model and accepts piece-wise constant inputs at the beginning of each step, is used in all
    simulation. The robot's initial pose and position of the final goal were hand selected. The highlighted areas show the local maps being provided to the robot $ 8 \times 8$ for left and middle columns and $9 \times 9$ for the right column. In each case, the planner guided the robot to the goal. Animations of the simulations are available at \cite{githubFileCLFPlanning}.}
\label{fig:FakeMap}%
\end{figure*}

\subsubsection{Sub-goal Finder and Finite-State Machine}
Ideally, a global planner is present to guide the robot to a distant goal,
which may not be viewable at the time of mission start \cite{blochliger2018topomap}. In relatively simple situations such as that shown in
Fig.~\ref{fig:FakeMap} and Fig.~\ref{fig:ZoomedLocalMap}, it is sufficient to complete many of short-term
missions by positioning a sub-goal (green arrow) at the lowest cost (cost-to-come +
cost-to-goal) on an arc (blue arrows) to guide the robot to the final goal. This
sub-goal finder is also used as a finite-state machine to handle high-level missions
such as making turn selections at intersections. In the future, the sub-goal finder will
be replaced with a global planner.

\subsection{Reactive Thread}
\label{sec:ExecutionThread}

The work in \cite{golbol2018rg, arslan2019sensor, paternain2017navigation, ExactRobot2016, koditschek1990robot, rimon1990exact, borenstein1989real, koditschek1987exact, gomez2013planning} provided a significant alternative to the standard path tracking. Their \textit{high frequency reactive planners} create a vector field on the configuration space whose integrals curves (i.e., solutions of the vector field) provide alternative paths to the goal. When the robot is perturbed, it immediately starts following the new path specified by the vector field, instead trying to asymptotically rejoin the original path. The vector field is in essence an instantaneous re-planner.

In the reactive planner of \cite{arslan2019sensor, ExactRobot2016}, the vector field arises from the gradient of a potential function defined on the configuration space. Here, we use the solutions of the closed-loop system associated with the CLF in \eqref{eq:CLFDistance} to define alternative paths in the configuration space. In essence, our feedback functions \eqref{eq:CLFVxVy} and \eqref{eq:CLFOmega} provide instantaneous re-planning of the control commands for the omnidirectional model \eqref{eq:RobotState}. This reactive planner can be run at 300 Hz in real-time.

The reactive thread is a reactive planner, in which the motion of the robot is
generated by a vector field that replies on a closed-loop feedback policy giving
controller commands in real-time as a function of the instantaneous robot pose. In
other words, the reactive planner utilizes the proposed Control Lyapunov Function
described in Sec.~\ref{sec:CLF} to adjust controller commands automatically when the
robot deviates from the optimal path. This thread steers the robot to the optimal
path at 300 Hz.

\begin{remark}
The ``timing'' of Cassie's foot placement is inherently event-driven and stochastic. Even though a step cycle may be planned for 300 ms, variations in the terrain and deviations of the robot's joints from nominal conditions result in foot-ground contact being a random variable, with a mean of roughly 300 ms. Running the reactive planner at anything over 100 Hz essentially guarantees that Cassie's gait controller, which runs at 2 kHz, is accepting the most up-to-date commands from the planner, even if a few messages are lost over UDP transmission.
\end{remark}

\section{Simulation and Experimental Results}
\label{sec:Experiment}
The proposed reactive planning system integrates a local map, the omnidirectional
CLF \rrtN, and fast replanning from the reactive thread. We performed three types of
evaluation of the reactive planning system.

\subsection{Angular Linear Inverted Pendulum (ALIP) Robot with Simulated Challenging Outdoor Terrains and Indoor Cluttered Scenes}
\label{sec:Simulation}
We first ran the reactive planning system on several synthetic environments, in which an
ALIP robot model \cite{131811, gong2020angular} navigated several simulated noisy, patchy, challenging outdoor terrains as well as cluttered
indoor scenes. The ALIP robot successfully reached all the goals in different
scenes. We tested the system on more than 10 different environments, both indoor and
outdoor with and without obstacles. Due to space limitations, we only show the results of 
six simulations in Fig~\ref{fig:FakeMap}; see our
GitHub\cite{githubFileCLFPlanning} for videos and more results. 

\begin{remark}
    The ALIP robot\cite{131811, gong2020angular} takes piece-wise constant inputs from the reactive planning
    system. Let $g, H, \tau$ be the gravity, the robot's center of mass  height, and the time interval
    of a swing phase, respectively. The motion of an ALIP robot on the \xaxis is
    defined as 
    \begin{equation}
        \begin{bmatrix}
            x_{k+1}\\
            \dot{x_{k+1}}
        \end{bmatrix} = 
        \begin{bmatrix}
            \cosh(\xi) & \frac{1}{\rho}\sinh(\xi)\\
            \rho\sinh(\xi) & \cosh(\xi)
        \end{bmatrix}
        \begin{bmatrix}
            x_{k}\\
            \dot{x_{k}}
        \end{bmatrix} +
        \begin{bmatrix}
            1-\cosh(\xi) \\
            -\rho\sinh(\xi)
        \end{bmatrix} p_x,
    \end{equation}
    where $p_x$ is the center of mass (CoM) on the \xaxis of the robot, $\xi =
    \rho\tau$ and $\rho = \sqrt{g/H}$. Similarly, the motion of the robot on the
    \yaxis can be defined.
\end{remark}

\begin{remark}
   Even though a full global map is given in each simulation environment, only the information in the local map is given to the planning system at each
   timestamp. The path generated from omnidirectional \rrt is asymptotically optimal within the local map, for the given time window. It is emphasized that no global information is provided to the planner which is why the resulting trajectory from the initial point to the goal may not be the shortest path. 
\end{remark}

\subsection{Validation of Control Command Feasibility via a Whole-body Cassie Simulator}
\begin{figure}[t]%
\centering
\includegraphics[width=1\columnwidth]{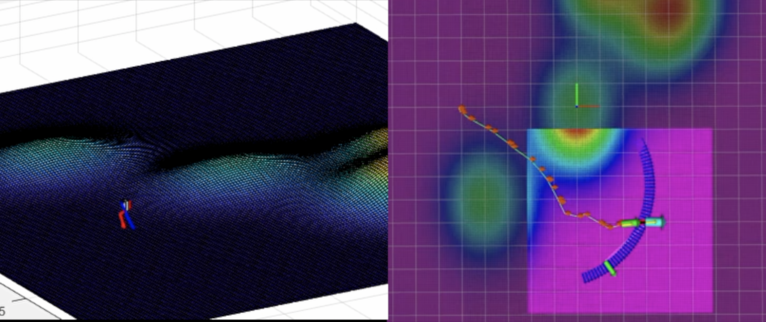}
\caption[]{A simulation of a C++-implementation of the reactive planner on full-dynamic model of Cassie, which accounts for all 20 degrees of freedom
of Cassie in Matlab-Simmechanics and a 3D terrain model. The reactive planning system receives the pose of the simulated Cassie via User Datagram Protocol (UDP). Cassie's simulator receives and executes the resulting control commands
via UDP. The planning system successfully takes the simulated
Cassie to the goal without falling. An animation is available at \cite{githubFileCLFPlanning}}
\label{fig:CassieSimulation}%
\end{figure}

To ensure the control commands from the reactive planning system are feasible for
Cassie-series bipedal robots, we sent the commands via User Datagram Protocol (UDP) from ROS\cite{ROS}
C++ to Matlab-Simmechanics, which simulates a 20 DoF of Cassie, using footfalls on the specified terrain. The simulator then sent back the pose of the simulated Cassie robot to the planning system to plan for
the optimal path via UDP. The planner system successfully took the simulated Cassie to the
goal without falling, as shown in Fig.~\ref{fig:CassieSimulation}.

\subsection{System Integration for Real-time Deployment}
\label{sec:SystemIntegration} 
\begin{figure}[t]%
\centering
\includegraphics[width=1\columnwidth]{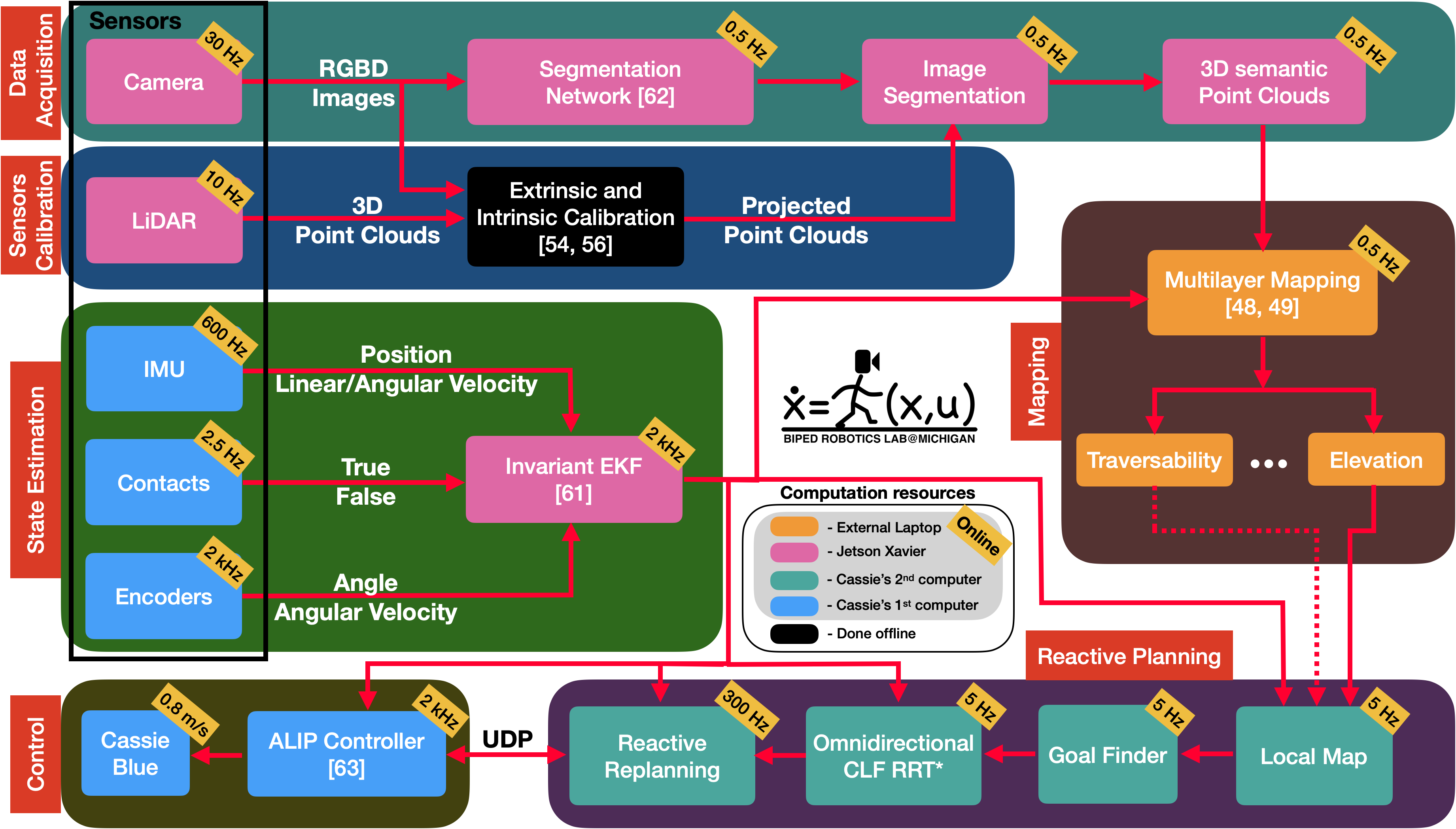}
\caption[]{Illustration of how the various processes in the overall autonomy system are distributed and their computation frequencies. The larger boxes indicate various modules such as Data Acquisition, Planning, and Control. The smaller boxes are colored according to the processor that runs them.
}
\label{fig:SystemIntegration}%
\end{figure}

System integration is critical for real-time use. 
Figure~\ref{fig:SystemIntegration} shows the integrated system, distribution, and
frequency of each computation. In particular, the sensor calibrations are performed
via\cite{rehder2016extending, furgale2013unified, oth2013rolling,
huang2020improvements, huang2021lidartag, huang2020intinsic,huang2020lidartaglonger, githubFileExtrinsic,
githubFileLiDARTag, githubFileIntrinsic}. The invariant Extended Kalman Filter 
(InEKF)\cite{Hartley-RSS-18, hartley2019contact} is used to estimate the state of
Cassie Blue at 2 kHz. Images are segmented via
MobileNets\cite{howard2017mobilenets} and a \lidar point cloud is projected back to
the segmented image to produce a 3D segmented point cloud. The
resulting point clouds are then utilized to build a multi-layer map
(MLM)~\cite{Fankhauser2014RobotCentricElevationMapping,
Fankhauser2018ProbabilisticTerrainMapping, Lu2020BKI}. 
The reactive planning system then crops the MLM around the robot position
to create a local map and performs several operations to acquire extra information, as
described in Sec.~\ref{sec:PlanningThread}. Additionally, the reactive planner receives the robot poses from the InEKF at 300 Hz to adjust the control commands that guide the
robot to the nominal sub-poses via the proposed CLF; see Sec.~\ref{sec:CLF} and
Sec.~\ref{sec:ExecutionThread}. The control commands are then sent to Cassie Blue's
gait controller~\cite{gong2020angular, gong2021zero, gong2019feedback} via UDP.


\subsection{Full Autonomy Experiments with Cassie Blue}
\label{sec:FullyAutonomyExps}
We conducted several indoor and outdoor full autonomy experiments with Cassie
Blue.
\subsubsection{The Wave Field}
\begin{figure}[t]%
\centering
\begin{subfigure}{0.5\columnwidth}
    \centering
\includegraphics[width=0.88\columnwidth]{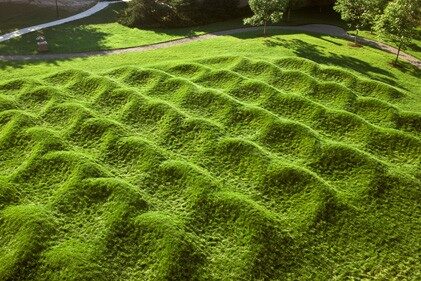}%
\caption{}
\label{fig:WaveFieldPic}
\end{subfigure}%
\begin{subfigure}{0.5\columnwidth}
    \centering
\includegraphics[trim={0cm 5cm 0cm 3cm},clip,width=0.95\columnwidth]{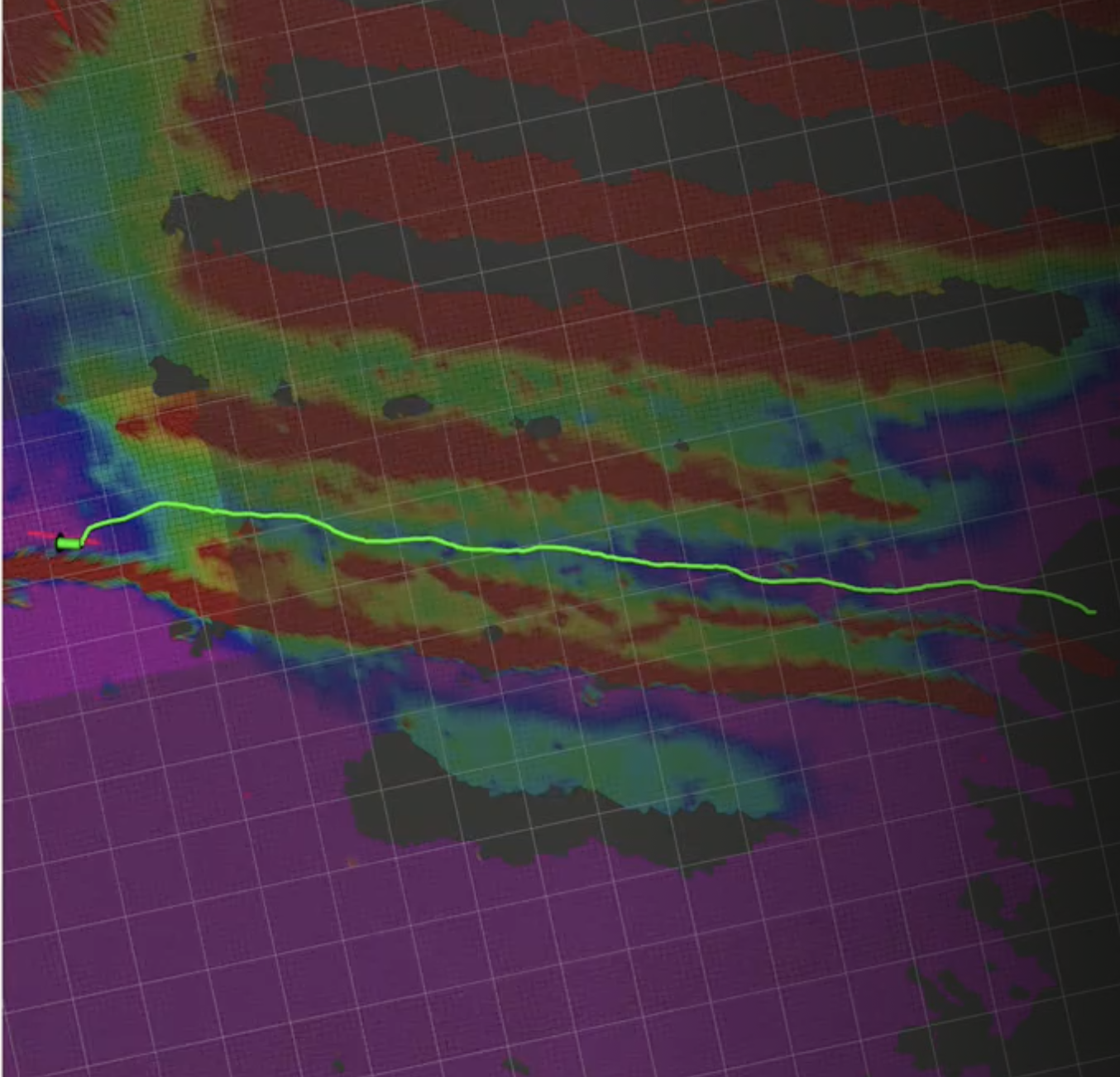}%
\caption{}
\label{fig:ExpFull}%
\end{subfigure}
\begin{subfigure}{0.5\columnwidth}
    \centering
\includegraphics[height=0.8\columnwidth]{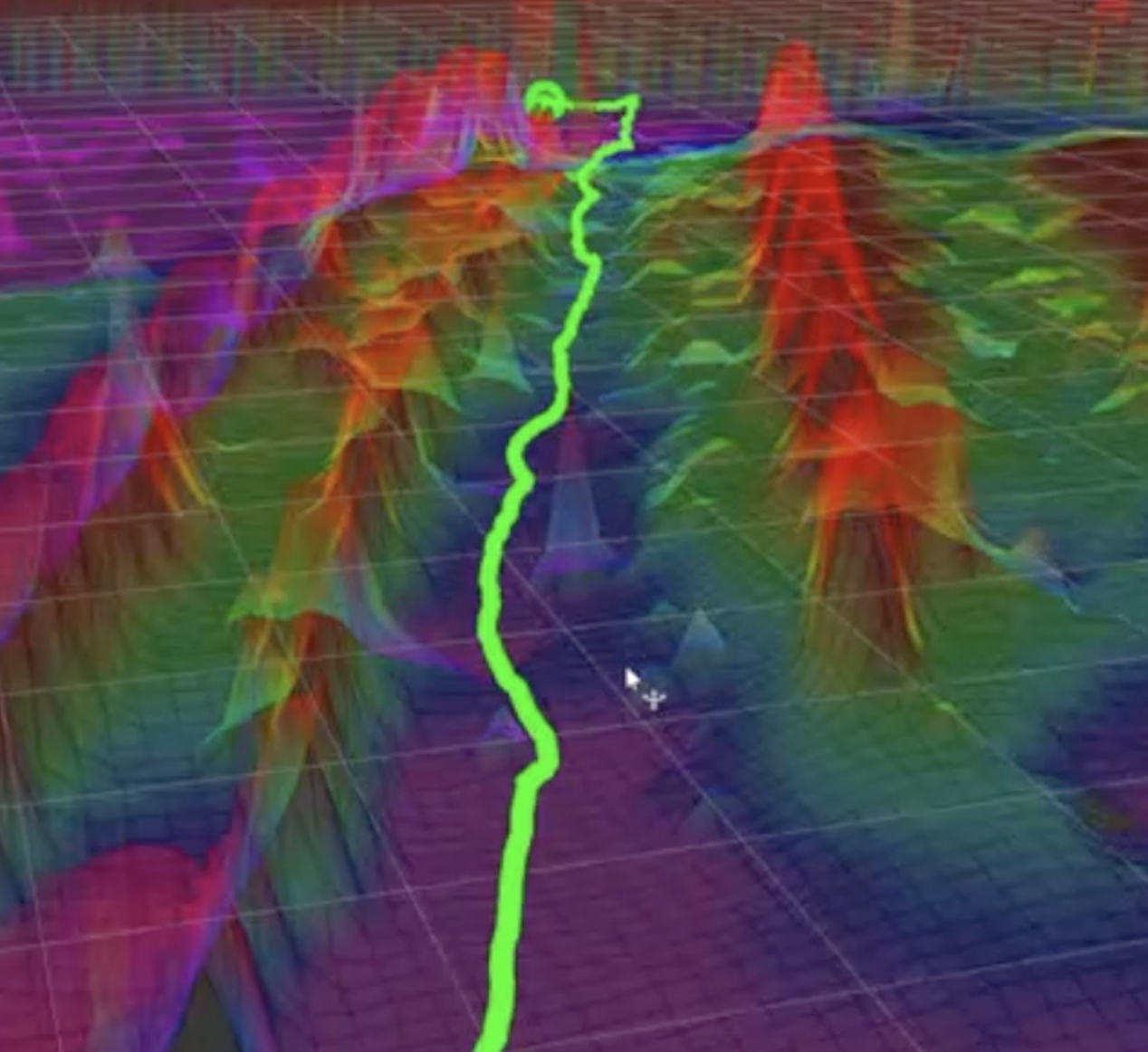}~
\caption{}
\label{fig:ExpBack}%
\end{subfigure}%
\begin{subfigure}{0.5\columnwidth}
    \centering
\includegraphics[height=0.8\columnwidth]{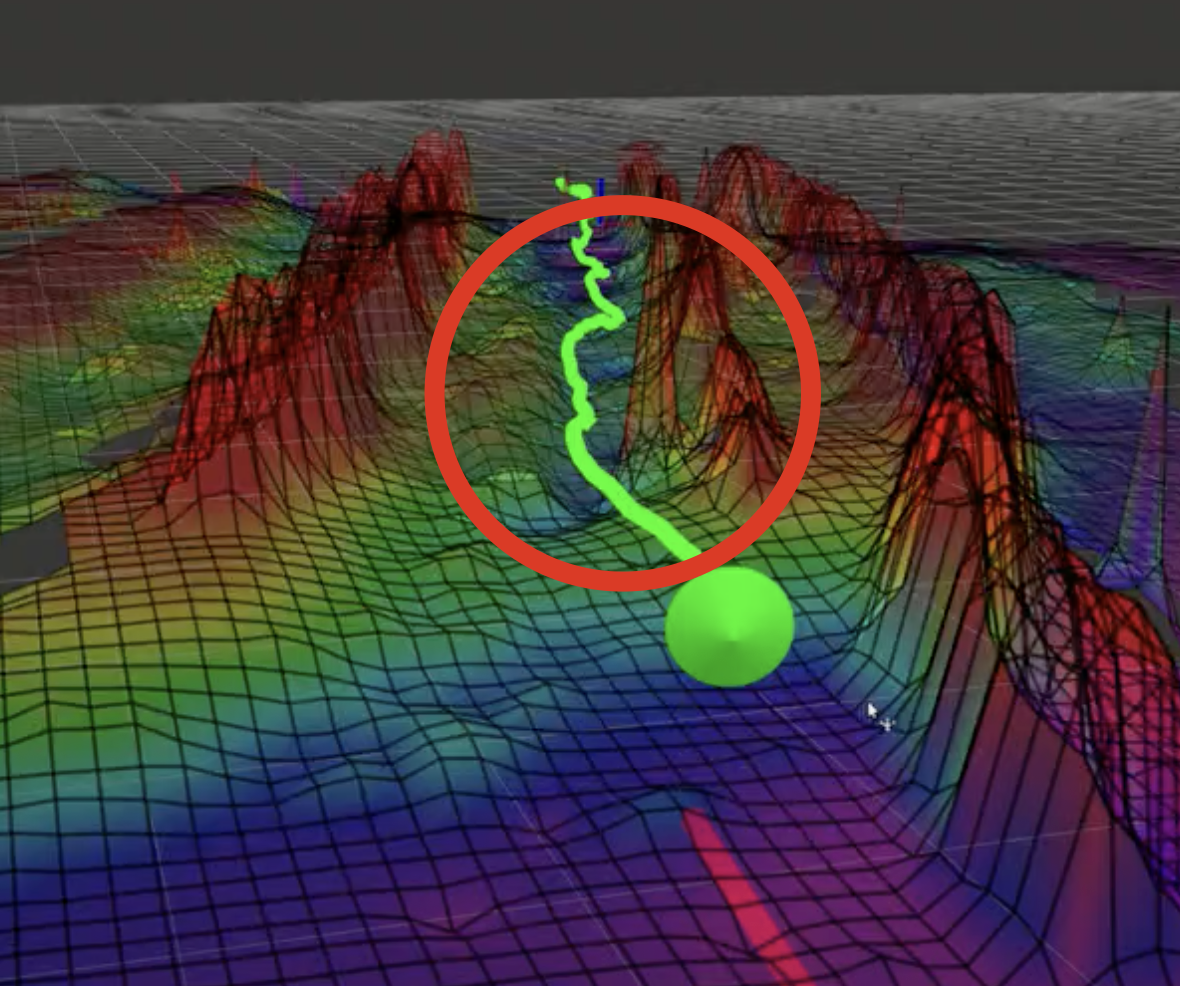}%
\caption{}
\label{fig:ExpObs}%
\end{subfigure}%
\caption[]{Experimental results on the Wave Field. The top-left shows the experiment terrain, the Wave Field, on the North Campus of the University of Michigan. The top-right shows a bird's-eye view of the
resulting trajectory from the reactive planning system. The bottom-left shows a
back-view of the trajectory produced by the planning system as Cassie Blue walks in a  valley (highly traversable area) of the Wave Field. The bottom-right demonstrates the planning system avoiding areas of higher cost.}%
\label{fig:ExpWaveField}%
\end{figure}

\begin{figure}[t]%
\centering
\includegraphics[width=1\columnwidth]{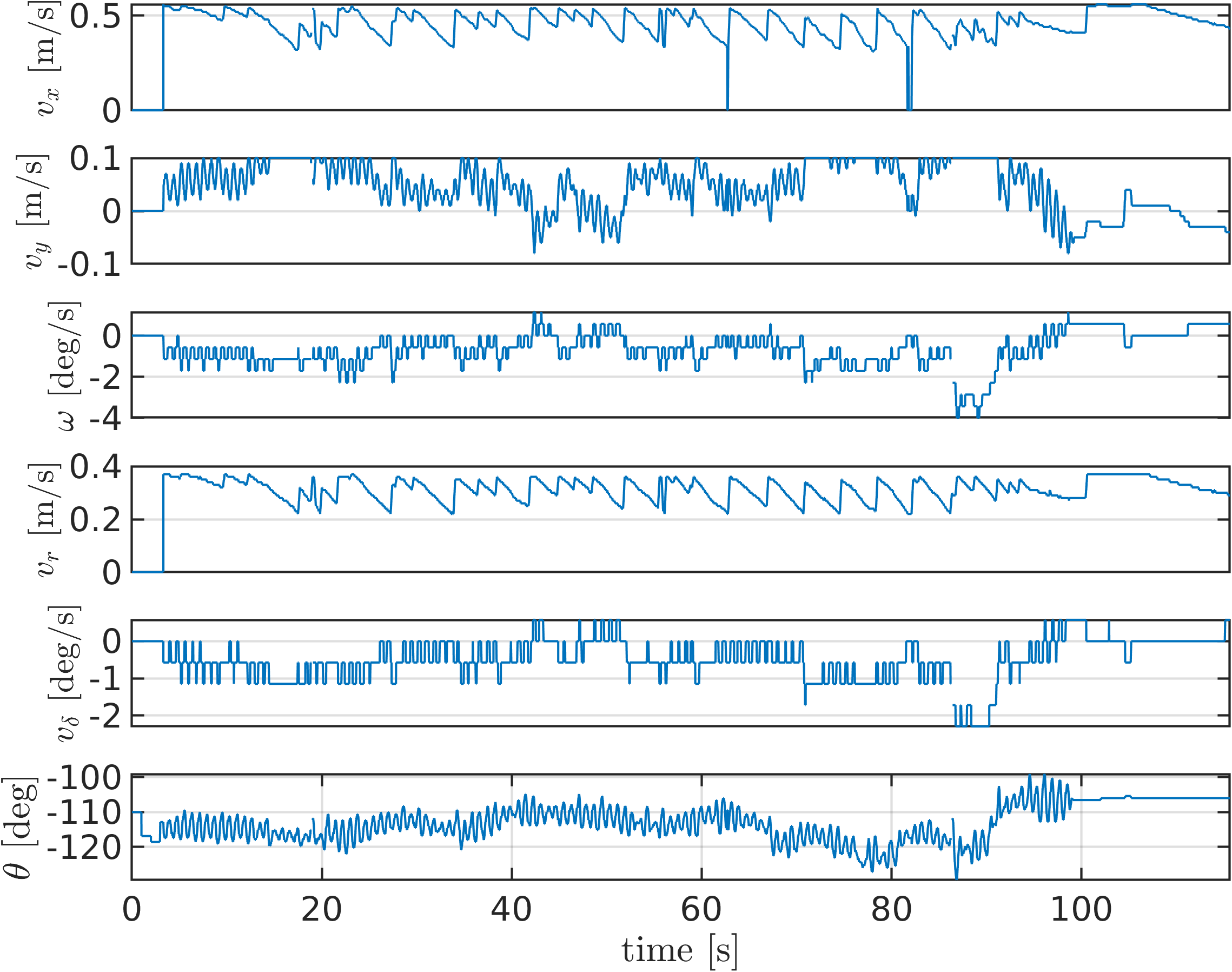}
\caption[]{Control Commands sent to Cassie Blue.}
\label{fig:ExpControlCommands}%
\end{figure}

We achieved full autonomy with Cassie Blue on the Wave Field, located on the North
campus of the University of Michigan, an earthen sculpture designed by Maya Lin
\cite{PhotoOfWaveField}; see Fig.~\ref{fig:WaveFieldPic}. The Wave Field consists of
sinusoidal humps with a depth of approximately 1.5 m from the bottom of the valleys
to the crest of the humps; there is a second sinusoidal pattern running orthogonal to
the main pattern, which adds 25 cm ripples peak-to-peak even in the valleys.
Figure~\ref{fig:ExpFull} shows the top-view of the resulting trajectory of the
reactive planning system. The planning system guided Cassie Blue to walk in the
valley (the more traversable area), as shown in Fig.~\ref{fig:ExpBack}. The planning
system navigated Cassie Blue around a hump that protrudes into one of the valleys, as
shown in Fig.~\ref{fig:ExpObs}. Figure~\ref{fig:ExpControlCommands} shows the control
commands sent to Cassie Blue. This experiment was presented in the Legged Robots
Workshop at ICRA 2021; the video can be viewed at \cite{Autonomy2021ICRAWorkshop}.
The video of the Wave Field experiment is uploaded and can be found at
\cite{WaveFieldAutonomy2021} and \cite{githubFileCLFPlanning}. 

\subsubsection{Turn left at detected intersections of corridors and avoid obstacles} 
\begin{figure}[t]%
\centering
\begin{subfigure}{1\columnwidth}
    \centering
\includegraphics[width=1\columnwidth]{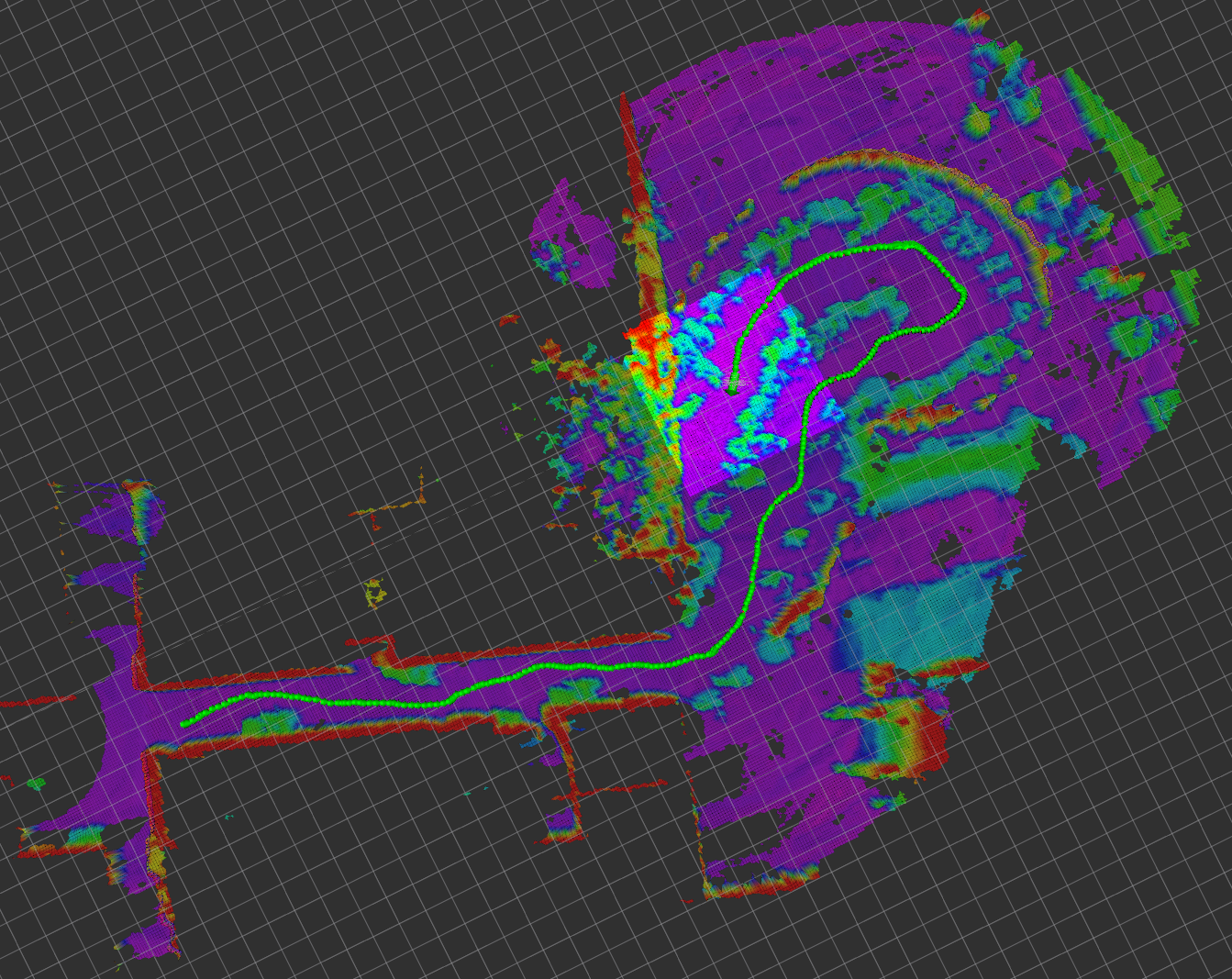}%
\label{fig:1stFloorFRB}
\end{subfigure}%
\vspace{2mm}
\begin{subfigure}{1\columnwidth}
    \centering
\includegraphics[width=1\columnwidth]{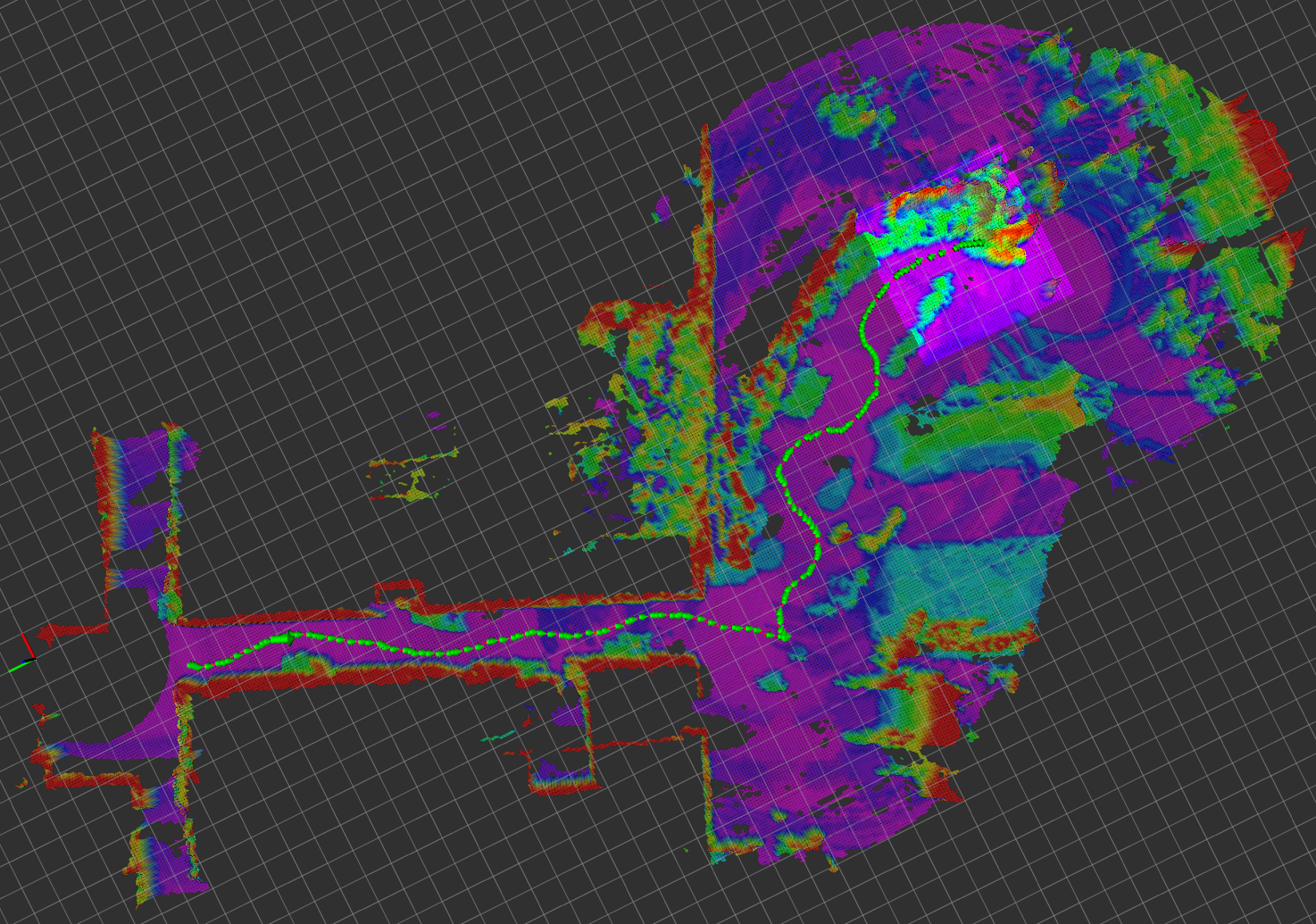}~
\label{fig:1stFloorExp}%
\end{subfigure}%
\caption[]{The resulting trajectories on the first floor of the Ford Robotics Building. 
    The map colored by height was built online while Cassie was guided by the planning system. The
    green lines are the resulting trajectories and green patches in the map are
    tables and furniture considered obstacles.
}%
\label{fig:FirstFloorFRBExp}%
\end{figure}

We conducted two experiments of this type on the first floor of the Ford Robotics Building (FRB)
at the University of Michigan. The experiments' scenes consist of corridors and an open
area cluttered with tables and couches, which are considered as obstacles, as shown
in Fig.~\ref{fig:FirstFloorFRBExp}. To detect the intersections of the corridors, we
group walkable segments within a ring around Cassie Blue via the single-linkage
agglomerative hierarchical clustering algorithm\footnote{We chose this clustering
    algorithm because the number of clusters is unknown. Therefore, algorithms like
K-Means Clustering \cite{lloyd1982least} cannot be used.}\cite{johnson1967hierarchical}.
Subsequently, Cassie Blue makes a left turn at the detected intersection. After
exiting the corridors, the robot reaches an open area cluttered with furniture
and performs obstacle avoidance. Under the proposed reactive planning
system, Cassie Blue completed the experiments without falling or
colliding with obstacles. The total distance traveled was about 80 meters.
The experiment videos can be viewed at \cite{FRBAvoidance} and
\cite{githubFileCLFPlanning}.

\subsubsection{Turn right at detected intersections of corridors and return to the initial position} 
\begin{figure}[t]%
\centering
\begin{subfigure}{1\columnwidth}
    \centering
\includegraphics[width=1\columnwidth]{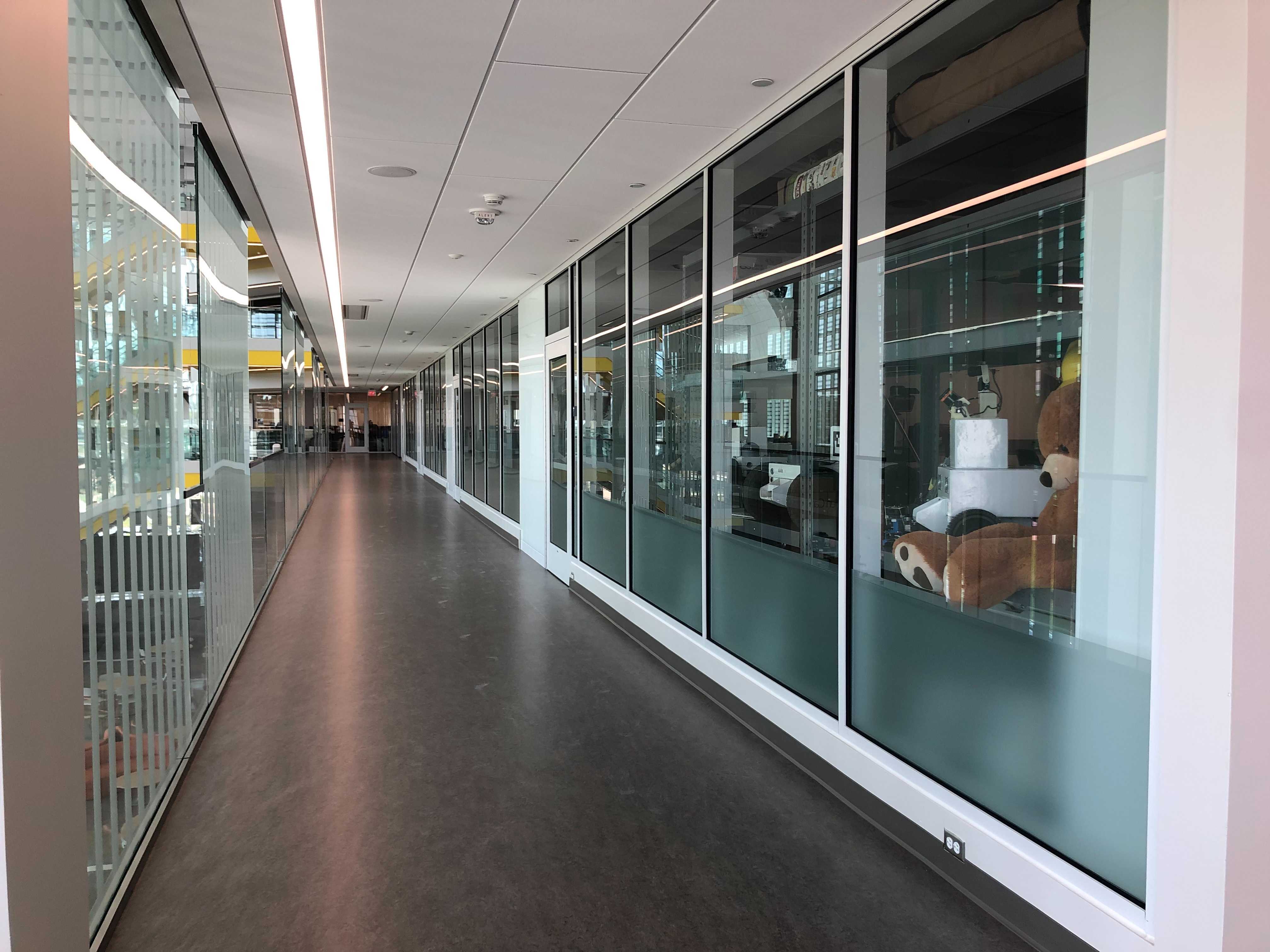}%
\label{fig:2ndFloorFRB}
\end{subfigure}%
\vspace{2mm}
\begin{subfigure}{1\columnwidth}
    \centering
\includegraphics[width=1\columnwidth]{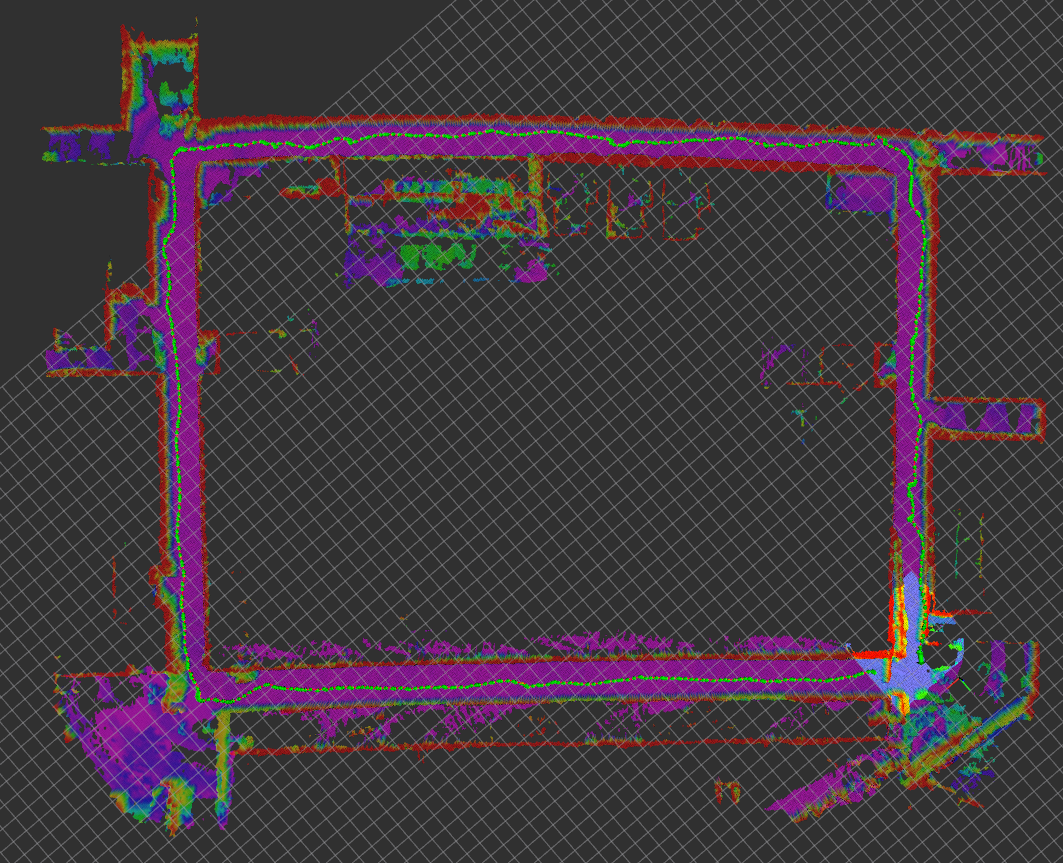}
\label{fig:2ndFloorExp}%
\end{subfigure}%
\caption[]{Experimental results on the second floor of the Ford Robotics Building. The top shows
    glass walls, which lead to refection of \lidar lasers and creating walkable area
    behind the wall. The bottom illustrates the resulting trajectory produced by the
    planning system as Cassie Blue walks. 
}%
\label{fig:SecondFloorFRBExp}%
\end{figure}

This experiment was conducted on the second floor of the FRB and the experiment scene
contains four long corridors with glass walls. Some of the \lidar beams penetrated glass a certain points along the corridors, causing the mapping algorithm to consider area behind the
glass walls as free and walkable. We applied the Bresenham line algorithm
\cite{bresenham1965algorithm} to remove the walkable area behind the glass walls.
The computation of the Bresenham algorithm is not expensive because it is only applied
within the local map, mentioned in Sec.~\ref{sec:PlanningThread}.
The proposed reactive planning system successfully guided Cassie Blue back to its
initial position, as shown in Fig.~\ref{fig:SecondFloorFRBExp}. The total distance traveled was about 200 meters. 
The experiment videos can be viewed at \cite{FRBReturnToOrigin} and
\cite{githubFileCLFPlanning}.

\subsection{Experiment Discussion}
In the two indoor experiments, Cassie exhibited a walk-and-stop motion. Where does it come from?
    As mentioned in Sec.~\ref{sec:PlannerSystem}, the planning threading runs at 5
    Hz. At the $k$-th update, there will be an optimal path $\Pcal_k$, comprised of a
    number of way-poses connected by CLFs. Although each vector field associated to a
    CLF is continuous (even smooth), switching among CLFs can induce discontinuity. This
    discontinuity induces Cassie's walk-and-stop motion seen in the videos of the indoor
    experiments. How? At each planning update, the entire tree was being discarded and a new one constructed. In particular, the closest way-pose to Cassie was being re-set every 200 ms, and thus the robot was never allowed to evolve along the integral curves of the vector field.
    
    The solution is straightforward: at the $(k\!\!+\!\!1)$-st planning update, we leave the first unreached way-pose fixed in the path $\Pcal_k$ to ensure continuity. Additionally,
to fully utilize the optimal path from the previous update, we keep the current optimal path
$\Pcal_k$ as a branch and prune all the samples from the $k$-th update. This provides 
a warm start for the $(k\!\!+\!\!1)$-st update, as long as the path $\Pcal_k$ is still
valid and collision-free. If a dynamic obstacle has invalidated the path between the robot's current position and the first unreached way-pose, then the entire tree is discarded, as before. With these changes made, we conducted several additional
experiments to confirm that it resolves the walk-and-stop movement. The experiments can be viewed at
\cite{FRBSmooth} and \cite{githubFileCLFPlanning}. 




\section{Conclusion and Future Work}
\label{sec:Conclusion}
We presented a novel reactive planning system that consists of a 5-Hz planning thread
to guide a robot to a distant goal and a 300-Hz CLF-based reactive thread to cope
with robot deviations. In simulation, we evaluated the reactive planning system on
ten challenging outdoor terrains and cluttered indoor scenes. In experiments on Cassie Blue, a bipedal robot with 20 DoF, we performed fully autonomous navigation outdoors on sinusoidally varying terrain and indoors in cluttered hallways and an atrium.

The planning thread uses a multi-layer, robot-centric local map to compute
traversability for challenging terrains, a sub-goal finder, and a finite-state machine to
choose a sub-goal location as well as omnidirectional CLF \rrt to find an
asymptotically optimal path for Cassie to walk in a traversable area. The
omnidirectional CLF \rrt utilizes the newly proposed Control-Lyapunov function (CLF)
as the steering function and the distance measure on the CLF manifold in the \rrt
algorithm. Both the proposed CLF and the distance measure have a closed-form
solution. The distance measure nicely accounts for the inherent ``features'' of
Cassie-series robots, such as high-cost for lateral movement. The robot's motion in the
reactive thread is generated by a vector field depending on a closed-loop feedback
policy providing control commands to the robot in real-time as a function of
instantaneous robot pose. In this manner, problems typically encountered by
waypoint-following and pathway-tracking strategies when transitioning between
waypoints or pathways (unsmooth motion, sudden turning, and abrupt acceleration) are resolved.
 
In the future, we shall combine control barrier functions \cite{7782377, 7040372,
7864310, 7798370, xu2015robustness, nguyen2020dynamic} with the CLF in the reactive
thread to handle dynamic obstacles. Additionally, the current local map is a 2.5D,
multi-layer grid map with fixed resolution; it is also interesting to see how to
efficiently represent a continuous local map. Furthermore, how to extend the CLF to
3D is another interesting area for future research.


\section*{Acknowledgment}
\small{ 
Toyota Research Institute provided funds to support this work. Funding for J. Grizzle
was in part provided by NSF Award No.~1808051 and 2118818. This article solely reflects the
opinions and conclusions of its authors and not the funding entities. The authors thank Lu Gan and Ray Zhang for their assistance in the development of the autonomy package used on Cassie Blue in these experiments and Yukai Gong and Dianhao Chen for the low-level gait controller used in Cassie. They also thank Dianhao Chen, Jinze Liu, Jenny Tan, Dongmyeong Lee, Jianyang Tang, and Peter
Wrobel, Minzhe Li, Lu Gan, Ray Zhang, Yukai Gong, and Oluwami Dosunmu-Ogunbi for their assistance in the experiments. The first
author thanks to Jong Jin Park, Collin Johnson, Peter Gaskell, and Prof.
Benjamin Kuipers for kindly providing insightful discussion for their work.
The first author thanks Wonhui Kim for useful
conversations.} 

\balance
\bibliographystyle{DefinesBib/bib_all/IEEEtran}
\bibliography{DefinesBib/bib_all/strings-abrv,DefinesBib/bib_all/ieee-abrv,DefinesBib/bib_all/BipedLab.bib,DefinesBib/bib_all/Books.bib,DefinesBib/bib_all/Bruce.bib,DefinesBib/bib_all/ComputerVision.bib,DefinesBib/bib_all/ComputerVisionNN.bib,DefinesBib/bib_all/IntrinsicCal.bib,DefinesBib/bib_all/L2C.bib,DefinesBib/bib_all/LibsNSoftwares.bib,DefinesBib/bib_all/ML.bib,DefinesBib/bib_all/OptimizationNMath.bib,DefinesBib/bib_all/Other.bib,DefinesBib/bib_all/StateEstimationSLAM.bib,DefinesBib/bib_all/MotionPlanning.bib,DefinesBib/bib_all/Mapping.bib,DefinesBib/bib_all/TrajectoriesOptimization.bib,DefinesBib/bib_all/Controls.bib}
\end{document}